\documentclass[twocolumn,pre,nofootinbib]{revtex4}   

\usepackage{dcolumn}
\usepackage{amsmath}
\usepackage{newtxtext,newtxmath}
\usepackage{rotating}
\usepackage{multirow}
\usepackage{graphicx}
\usepackage{subfigure}

\newcommand{\etal}{{\it{}et~al.}}

\newcommand{\fl}{\ensuremath{\flat}}
\newcommand{\sh}{\ensuremath{\sharp}}
\newcommand{\goesto}{\ensuremath{\rightarrow}}
\newcommand{\argmax}{\mathop\mathrm{argmax}}
\newcommand{\Ord}{\mathrm{O}}
\newcommand{\cC}{\mathcal{C}}
\newcommand{\cK}{\mathcal{K}}
\newcommand{\dm}{o}
\newcommand{\hdm}{\o}
\newcommand{\s}[3]{$\textrm{#1}^{\textrm{#2}}_{\textrm{#3}}$}

\begin{document}

\title{Computational music analysis from first principles}
\author{Dmitri Tymoczko$^{1,3}$ and Mark Newman$^{2,3}$}
\affiliation{$^1$Department of Music, Princeton University, Princeton, NJ 19147, USA}
\affiliation{$^2$Center for the Study of Complex Systems, University of Michigan, Ann Arbor, MI 48109, USA}
\affiliation{$^3$Santa Fe Institute, 1399 Hyde Park Road, Santa Fe, NM 87501, USA}

\begin{abstract}
We use coupled hidden Markov models to automatically annotate the 371 Bach chorales in the Riemenschneider edition, a corpus containing approximately 100\,000 notes and 20\,000 chords.  We give three separate analyses that achieve progressively greater accuracy at the cost of making increasingly strong assumptions about musical syntax.  Although our method makes almost no use of human input, we are able to identify both chords and keys with an accuracy of 85\% or greater when compared to an expert human analysis, resulting in annotations accurate enough to be used for a range of music-theoretical purposes, while also being free of subjective human judgments.  Our work bears on longstanding debates about the objective reality of the structures postulated by standard Western harmonic theory, as well as on specific questions about the nature of Western harmonic syntax. 
\end{abstract}
\maketitle

\section{Introduction}
\begin{figure}[b]
\begin{center}
\includegraphics[width=\columnwidth]{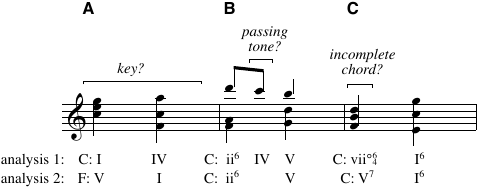}
\end{center}
\caption{Three sources of ambiguity in harmonic analysis.  (A)~These chords can be heard in C major or F major.  (B)~We can consider the note C as a nonharmonic passing tone, or as forming an F major chord.  (C)~We can consider the first chord to be an incomplete G7 chord missing its root.}
\label{fig:ambiguity}
\end{figure}

\begin{figure}[b]
\begin{center}
\includegraphics[width=2.325in]{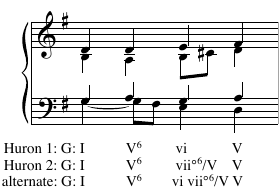}
\end{center}
\caption{Three possible analyses of a single passage.  David Huron's analyses assign a single chord to the third quarter note.  For reasons discussed in Section \ref{sec:improvements} we prefer an analysis with separate chords on each eighth note.}
\label{fig:huron}
\end{figure}

More than two hundred years ago, Gottfried Weber argued that traditional Western harmony was pervasively ambiguous~\cite{WEBER1846}.   One problem involves chords and keys: a single progression such as C major \goesto\ F major can be experienced as a tension-reducing dominant-tonic in F major or a tension-increasing tonic-subdominant in C major (Fig.~\ref{fig:ambiguity}).  A second problem involves chords themselves: a voice moving from D6 to C6 over a stable \{F4, A4\} can signify either a change of harmony, from an inverted D minor to a root-position F major, or a ``merely linear'' passing motion that creates an F major chord as a nonsyntactical byproduct.  A third problem involves incomplete sonorities, for instance bare thirds that could plausibly belong to multiple triads, or triads belonging to multiple seventh chords.  Together these ambiguities ensure that even relatively simple passages can be analyzed in multiple ways, with no agreed-upon rules for choosing among them.  Music analysis thus requires complex holistic judgments, balancing vague and often-conflicting heuristics \cite[pp.~303--310]{TAOM}.  It is not surprising, therefore, that we find widespread disagreement about analyses even among human experts---not only about general principles, but also specific passages (Fig.~\ref{fig:huron}).  

This holism helps explain the somewhat disappointing performance of unsupervised computational musical analysis.  Micchi et~al. \cite{Micchi}, for example, review a variety of attempts at automated analysis, none of which were able to assign Roman numerals with greater than 50\% overall accuracy.  In a different vein, Quinn and White~\cite{QUINNWHITE} used hidden Markov models to classify progressions in the Bach chorales but their results depart substantially from standard music-theoretical practice, for instance by including seemingly ``nonharmonic'' tones within chords, and by assigning tonic and subdominant to the same harmonic category.  Currently, no existing studies have managed to use unsupervised methods to  annotate substantial musical corpora in a way that closely captures the judgments of human annotators.  Some approaches have achieved a good match to human intuition in more limited contexts, such as the studies of Noland and Sandler~\cite{NOLAND2006} and N\'apoles L\'opez et al.~\cite{NAPOLESLOPEZ2019}, which both focused on the narrower question of identifying key centers rather than chords and were able to do so with about 90\% accuracy.

Substantial human disagreement, coupled with the lack\-luster performance of computational methods, together raise the skeptical possibility that human judgments might not be reducible to objective algorithms.  Weber himself argued that musical analysis required intuitive human judgment \cite[p.~658]{WEBER1846}.  Quinn~\cite{QUINN} has argued that the notion of a chord is theoretically superfluous, suggesting that one can accurately capture the syntax of traditional Western tonal music without distinguishing harmonic from nonharmonic tones.  In a similar vein, Gjerdingen~\cite{GJERDINGEN2007} proposes that the syntax of Western music is grounded in ``schemas'' that do not not require the apparatus of standard theory.  Perhaps the familiar theoretical picture, in which nonharmonic tones decorate tertian harmonies, is not so much an empirical claim as an interpretive grid we impose onto music.

Against this background, we present in this paper a simple and largely assumption-free computational method that does a good job at reproducing expert human analysis, achieving about 88\% overall correctness at the task of identifying keys and chords in the 371 Bach chorales of the Riemenschneider edition, compared to a human analysis of the same corpus.  Our approach encompasses a family of related algorithms all sharing the same basic structure: for a given input, they find a fixed number of transposable probability distributions that best represent the occurrence of elements (notes or chords) in that input.  When fed raw notes and told to find a small number of probability distributions, the output can be interpreted as a series of chords; when fed these same chords as input, the output can be interpreted as a series of keys.  The significance of these results lies not so much in their accuracy alone but in the combination of accuracy, simplicity, and lack of presupposition: our algorithm performs well with very little information about musical structure, using well-known and straightforward techniques, and without the need for extensive training sets of expert human analysis.  In using familiar methods to reproduce the standard chord-and-key picture, it answers skeptics who have questioned the objectivity of Western musical concepts \cite{QUINN}.  Conversely, the close correspondence to human annotations suggests that those annotations do not  rest on large numbers of arbitrary decisions.  Finally, by retracing the steps required to approximate human analyses, our algorithm clarifies the assumptions underlying traditional analytical practices.

The structure of the paper is as follows.  Section~\ref{sec:model} describes the statistical model we use in detail and places it in the context of previous work on related models.  Section~\ref{sec:results} applies this model to the corpus of Bach chorales and compares the results to those of a human analysis, finding reasonably good agreement.  Section~\ref{sec:improvements} improves these already-good results by applying a few additional rules of thumb concerning modulatory tendencies, local harmonic ``laws,'' and the principles of nonharmonic tone usage.  Section~\ref{sec:discussion} considers some broader methodological issues arising from our investigation, as well as potential uses for the machine-annotated corpus generated by our calculations.  A number of technical issues are discussed in appendices.  Data and code are freely available online at \verb|https://dmitri.mycpanel.princeton.edu/code.zip|.

\section{Hidden Markov models for analyzing\\
keys and chords}
\label{sec:model}
Our basic goal is to take a sequence of notes, represented as notation and divided among some number of voices, and infer from them the underlying chords and keys at each point in a work.  We ignore lengths of notes, rests, and other rhythmic elements of the music: all notes are given equal weight, whether short or long, and whether they fall on the beat or off it.  Whenever any voice begins a note or rest, we consider it a new data point for our analysis.  The key or chord need not change at every such data point however.

To perform our analysis we make use of hidden Markov models, a standard statistical tool for the analysis of discrete time-series data.  In a hidden Markov model it is assumed that a system of interest is governed by ``hidden'' factors, changing over time, which are not directly observed but which nonetheless influence the observed behavior of the system.  For example, the observed behavior might be the sung notes of a Bach chorale and the hidden factors the underlying chords, which one can model (for example) with Roman numerals.  The chords are hidden in the sense that the Roman numerals are not stated explicitly by the composer or written on the score, but they determine the observed data---the notes---in a probabilistic fashion.  If the underlying chord is C~major, for example, then there is a high probability that we will hear certain notes \{C,~E,~G\}, a~lower probability that we will hear some others \{D,~A,~B\}, and a still-lower probability for others \{C\sh,~E\fl,~A\fl\}.

The hidden factors can change over time.  Our C~major chord might transition to G~major, then to A~minor, and so forth, but some transitions are more common than others: C~major is unlikely to be followed by C\sh\ major for example.  To capture these differences, a hidden Markov model incorporates a set of transition probabilities that represent how likely every chord is to be followed by every other, and we expect that some of these probabilities will be  large and others small.

Collectively, these elements define the hidden Markov model: chords change over time with some transitions being more likely than others, and notes reflect the current chord in a probabilistic fashion, the probabilities of individual notes being dependent on the underlying chord.

The hidden Markov model can be thought of as a machine for generating sequences of notes from underlying chords.  In the present instance, however, we want to use the model in reverse, we want to \emph{infer} the chords from the notes.  We ask, ``If these are the notes we hear, what is our best guess about the chords that produced them?''  We can answer this question by combining the hidden Markov model with standard numerical techniques of Bayesian inference: the Baum-Welch and Viterbi algorithms, described below.

A central goal of our work is to perform these calculations while making as few assumptions as possible ahead of time.  We could for instance make use of guesses about the model parameters based on input from human experts, and analyses of this kind have been performed in the past~\cite{NOLAND2006,NAPOLESLOPEZ2019,BURGOYNE2005}, but we will not do that here.  The parameters in our calculations are given random initial values, which are then automatically refined by the Baum-Welch algorithm to fit the data.  We give the calculation as little help as possible in performing its analysis; we are interested in what it comes up with when given free rein to take any approach it likes.

Once we have inferred a collection of chords we use them as input to a second layer of analysis in which we infer the prevailing key at each moment from the chords.  This second analysis is also performed using a hidden Markov model.  The hidden features of this model are now the keys and it is assumed that chords are produced with probabilities that depend on the current key.  If we are in C~major for instance, then some chords will have a high probability of occurrence (C~major, G7), while others will have lower probability (C\sh~minor, A\fl~major).  Again we make only the most basic prior assumptions about the hidden Markov model and we initialize the parameter values at random; we are interested to see what the algorithm comes up with on its own.  In effect, the computer ``listens'' to a large corpus and ``learns'' how music works from first principles, based on what it hears.

Our only assumptions about our models are the following:
\begin{enumerate}
\item Music is built from an ordered set of 12 octave-free pitch classes, laid out in a circular arrangement like the numbers on a clock. Musical transposition is represented by addition modulo 12.
\item The probability of a particular note at a particular time is dictated by an underlying ``chord,'' a transposable probability distribution anchored on one of the 12 tones, and belonging to a small number of types, which is fixed at three in our calculations.  In practice these types turn out to correspond to accepted classes such as major and minor chords, but we do not specify this: we specify only the number of chord types and not their structure.
\item The probability of transition from one chord to the next depends on chord type and the interval by which the chords are separated.  We do not specify what the probabilities are, only that different probabilities can exist for different transitions.
\item The probabilities of appearance of chords are dictated by an underlying ``key,'' which is another transposable probability distribution anchored on one of the 12 tones, and belonging to a small number of types, fixed at two in our calculations.  In practice these types turn out to correspond to major and minor keys, but again we do not specify this, only that different types exist.
\item The probabilities of transition between keys may vary depending on the type of key and the interval between successive keys, but we do not specify what those probabilities are.
\end{enumerate}

Arguably the biggest assumption here is about the number of chord and key categories (three and two respectively), and these choices are certainly rooted in our pre-existing knowledge about the style.  The choice of two key categories, in particular, is based on the expectation that we will find major and minor keys only.  Other styles could employ a larger number of modes and demand a model with more key types.

The number of chords is more flexible.  Three categories gives reasonable results without artificially shoehorning highly dissimilar chords into a single category.  But we have performed analyses with larger numbers of chord types and the results are reasonable.  One can gain additional nuance by allowing more types, at the expense of more complex analytic output and also longer computation time.  Using three chord types, however, gives excellent results, so a larger number, while allowable, is not necessary.

\subsection{Related previous work}
A number of previous authors have employed hidden Markov models for harmonic analysis, particularly for the task of inferring the key of a piece of music, a calculation similar to the second half of our analysis described above.  The goal of this work has generally been to create machinery that can identify keys with high accuracy.  To do this, most authors have tuned their algorithms using knowledge from human experts concerning occurrence of notes within keys and transitions between keys.  While this can indeed produce high accuracy, it conflicts with our goal of determining how much the model can achieve on its own.  The difference of approach is analogous in some ways to different philosophies in the design of programs for playing chess.  Early chess programs were trained with large amounts of expert human knowledge about the game, such as opening books and endgames, and these programs can perform well.  But more recent work focused on a new class of algorithms, exemplified by DeepMind's Alpha\-Zero, which eschews human expertise and learns from scratch by playing against itself.  Starting with only the basic rules of the game, such as how the pieces move and what constitutes a win, the program is able in less than 24 hours to graduate from beginner chess to superhuman skill without outside help.

Most previous work on hidden Markov models for musical analysis has been analogous to the first approach, making use of human knowledge to build a working machine.  Noland and Sandler~\cite{NOLAND2006}, for example, made use of a hidden Markov model to learn keys from chord sequences in music by the Beatles, attaining 91\% accuracy in key identification compared to a human analysis.  The parameters of the model were initialized by hand using values assigned by experts.  N\'apoles L\'opez et~al.~\cite{NAPOLESLOPEZ2019} took a similar approach, but inferred keys directly from notes without using chords as input.  Their approach works best with symbolic input but can also be applied directly to audio data.  A number of other authors have also developed methods for analyzing audio data, most often in the form of pitch-class profiles~\cite{FUJISHIMA1999}.  Chai and Vercoe~\cite{CHAI2005} used this approach, for instance, to deduce keys directly from notes, while Sheh and Ellis~\cite{SHEH2003} and Burgoyne and Saul~\cite{BURGOYNE2005} both trained hidden Markov models to infer chords from pitch-class profiles.  Burgoyne and Saul went further to infer keys as well, although again using hand-tuned parameters.

Our approach makes use of no hand-tuned parameters---the only assumptions are those laid out in Section~\ref{sec:model}.  In effect we tell the program only the ``rules of chess'' and ask whether it can teach itself, from scratch, to interpret harmony as a human being does, simply by listening to enough of it.  We also use a larger corpus, annotated in much more detail, than the corpora featuring in previous work: our annotations record not only global keys but also local key changes, and include approximately 20\,000 chords.\footnote{Albrecht and Shanahan~\cite{ALBRECHTSHANAHAN} use a corpus of 982 pieces, but each is annotated only with a single global key; our corpus instead includes within-piece key changes.}  The corpus has been checked for accuracy both by hand and automatically, and it has been used in the past for a variety of purposes~\cite{TAOM}.

More recently, several authors have trained machine learning algorithms on corpora of hand-annotated scores \cite{napoleslopez21augmentednet}.  This strategy once again treats expert judgments as a capacity to be emulated.  Our work instead seeks to use machine learning to \emph{understand} that capacity, and in particular to determine the extent to which it is guided by arbitrary conventions.

\subsection{Inferring chords from notes}
\label{sec:chords}
We use two specific hidden Markov models in our calculations.  The first represents how notes are related to the underlying chords.  As applied to a complete corpus of musical works, such as a set of chorales, this model assumes a sequence of chords for each work, transition probabilities between chords, and note probabilities that depend on chords.  Chords are defined by an anchor pitch~$r$ and a type~$t$.  The anchor can be thought of roughly as the root of the chord, although in some cases, as we will see, it does not correspond to the conventional root.  Moreover, we are at pains to avoid any presuppositions about chord structure, so we use the generic term ``anchor'' here.

As discussed in Section~\ref{sec:model}, we fix the number of chord types at three.  We do not specify what the types represent---that is for the algorithm to learn---but as we will see they end up being major and minor triads plus a combined third type that captures both diminished and dominant chords.  For convenience we will often denote a chord by a single symbol~$c$, which is synonymous with the anchor/type pair~$(r,t)$.

As also mentioned previously, we identify a new data point every time any voice starts a note or rest.  We will refer to these data points as \emph{steps} of the work.\footnote{Quinn and White~\cite{QUINNWHITE} call them ``salami slices.''} Different steps do not necessarily correspond to the same note value---some could be quarter notes while others are eighth notes for instance.  The underlying chord may change at any step, or it may stay the same.  All chords are considered \textit{a~priori} equally likely at the start of a work and we define a set of probabilities~$\alpha_{c_1c_2}$ for subsequent transitions between chords $c_1=(r_1,t_1)$ and $c_2=(r_2,t_2)$ thus:
\begin{equation}
\alpha_{c_1c_2} = P( c_2 | c_1 ) = f(r_2-r_1,t_1,t_2).
\end{equation}
That is, the probability that the next chord will be $c_2$ given that the previous one was $c_1$ is a function of the types of the chords and the interval between their anchors.  Here $r_2-r_1$ indicates the upward interval from the first anchor to the second, in semitones modulo~12.  Thus the interval from B3 to C4 is 1~semitone while the interval from C4 to B3 is 11 semitones.  The probability of \emph{not} changing chord at a particular step is
\begin{equation}
\alpha_{c_1c_1} = f(0,t_1,t_1).
\end{equation}
Note that this does not depend on the anchor~$r_1$ of the chord but does depend on the type~$t_1$ and hence is potentially different for chords of different type.

We also define a set of probabilities for pitch classes to occur given the current chord.  Since pitch classes are octave free there are only 12 possibilities, which we represent by integers $n = 0\ldots11$.  Then
\begin{equation}
\mu_{cn} = P( n | c )
\end{equation}
is the probability that a sounding note has pitch~$n$ given that the current chord is~$c$.

This fully defines our first model.  Our goal is now to take the notes of all works in a corpus and fit the model to them, meaning we want to infer the chords and probabilities given the observed notes.  We fit a single instance of the model to all works in the corpus, under the assumption that similar rules of harmony apply to all works.  There are standard techniques for performing this type of fit, which comprise two stages of analysis~\cite{ZUCCHINI2021}.  In the first stage we use the Baum-Welch expectation-maximization algorithm to calculate the values of the probabilities.  Then in the second stage, given those values, we use the Viterbi dynamic-programming algorithm to calculate the best estimate of the chords.  A technical description of these two well-known algorithms is given in the appendices.

The Baum-Welch algorithm is an iterative algorithm that starts from an initial guess about the values of the parameters and repeatedly refines that guess until a locally optimal value is attained.  In previous work, authors have used input from human experts to make the initial guess, but in order to avoid relying on expert opinion we use random values produced by a high-quality random number generator.  Like all expectation-maximization algorithms, it is known that the Baum-Welch algorithm can converge to sub-optimal solutions from certain initial conditions.  To guard against this we perform 50 separate runs of the algorithm with different random starting conditions and take as our final solution the one that gives the highest value of the data likelihood---see Section~\ref{sec:results} and Appendix~\ref{sec:runs}.

Note that, since our calculations are performed using pitch classes as input, the chords produced by our calculation do not include information about inversions.  In Section~\ref{sec:improvements} we discuss simple rules for deducing inversions automatically by referring back to the score.

\subsection{Inferring keys from chords}
\label{sec:keys}
Once we have determined the chords, we use our second hidden Markov model to also determine the prevailing key at each step.  This model assumes a sequence of keys, one at each step, transition probabilities between keys, and probabilities of chords given the prevailing key.  Keys are represented by an anchor~$s$ and a type~$u$, and again for brevity we will use a single symbol~$k$ to represent a key, which is equivalent to the anchor/type pair~$(s,u)$.  (Again, the anchor can be intuitively associated with a tonic, but we do not assume this to be the case.)  We allow two types of keys but do not stipulate what they should be, although as we will see they strongly resemble major and minor.

Transitions between keys are parameterized by probabilities
\begin{equation}
\beta_{k_1k_2} = P( k_2 | k_1 ) = g(s_2-s_1,u_1,u_2),
\end{equation}
which is a function of the types of the keys and of the interval between their anchors (measured again as the upward interval in semitones, modulo~12).  All keys are considered \textit{a~priori} equally likely at the start of a work and the probability of not changing key is
\begin{equation}
\beta_{k_1k_1} = g(0,u_1,u_1),
\end{equation}
which thus depends on the type of key (as it should) but not on the anchor.  The probabilities of chords are
\begin{equation}
\nu_{kc} = P( c | k ),
\end{equation}
which is the probability that a sounding chord is~$c$ given the prevailing key~$k$.

This second model is fitted to the chord sequences generated by the first model (Section~\ref{sec:chords}), one sequence for each work in the corpus, using the Baum-Welch algorithm to determine the probabilities and the Viterbi algorithm to determine the specific keys.  Again we fit a single instance of the model to all works in the corpus, and again the initial parameter values are generated randomly and we repeat the entire calculation 50 times to guard against the possibility of converging to a sub-optimal solution.

The end result of applying both models to the corpus is a set of keys and chords at every step in every work.  As a by-product we also derive the values of the probability parameters, which are of interest in their own right, as discussed in Section~\ref{sec:results}.

\section{Analysis of Bach chorales}
\label{sec:results}
Our analysis focuses on the 371 Bach chorales in the Riemenschneider edition.  As described in Section~\ref{sec:model}, our calculations are divided into two parts, the first focusing on the chords and the second on the keys.   Each of these is then further subdivided into the calculation of the probability parameters and the calculation of the chords and keys themselves.

Turning first to the chords, our algorithm looks for chords of three types, which are distinguished by the probability distributions of their notes.  Over the course of the 50 runs of the algorithm, the best fit we find to the data is one in which the three chord classes correspond to major, minor, and dominant seventh chords, as described below.  The next best fit has classes corresponding to major, minor, and dominant seventh with suspended fourth, but this fit is enormously worse than the first-place winner: the data likelihood---the probability of the notes given the inferred chord classes---is more than a million times worse.  This allows us to say with the some confidence that the major/minor/dominant seventh analysis is the best fit to the data in the present case.  (See Appendix~\ref{sec:runs} for further discussion of this point.)

\begin{figure}
\begin{center}
\includegraphics[width=\columnwidth]{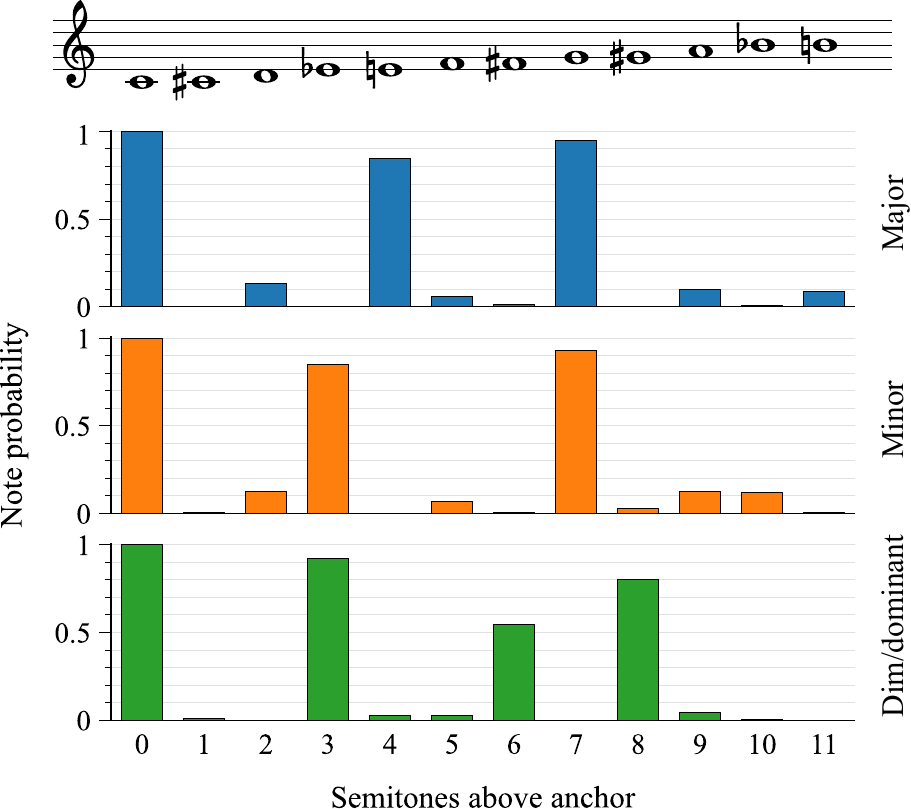}
\end{center}
\caption{Note probabilities for the three chord types that best capture the combinations commonly occurring in the corpus.  For each chord type we show the probability of occurrence of each of the 12 tones, relative to the anchor level, defined as the most likely note.  Notes relative to a C anchor are shown at the top for reference.  The three chord types closely match conventional major triads (top), minor triads (middle), and diminished triads/dominant sevenths (bottom).}
\label{fig:pnote}
\end{figure}

The distributions of notes within each chord class found by our method when applied to the chorales are shown in Fig.~\ref{fig:pnote}, with the convention that the highest-probability note is placed in the anchor position in each case.  The first two plots correspond clearly to major and minor chords---note the high probabilities assigned to the root, major third, and fifth in the first case and root, minor third, and fifth in the second case.  Other notes get much smaller probabilities, though nonzero, in part because these notes can occur as passing notes, neighbor notes, or other embellishments, and also because they occur as sevenths.  (See, for instance the minor seventh, 10 semitones above the anchor in the middle panel.  By the same token, the major sixth in the same panel, 9 semitones above the anchor, indicates the presence of a half-diminished seventh chord \{D, F, A, B\}.)

\begin{figure*}
\begin{center}
\includegraphics[width=\textwidth]{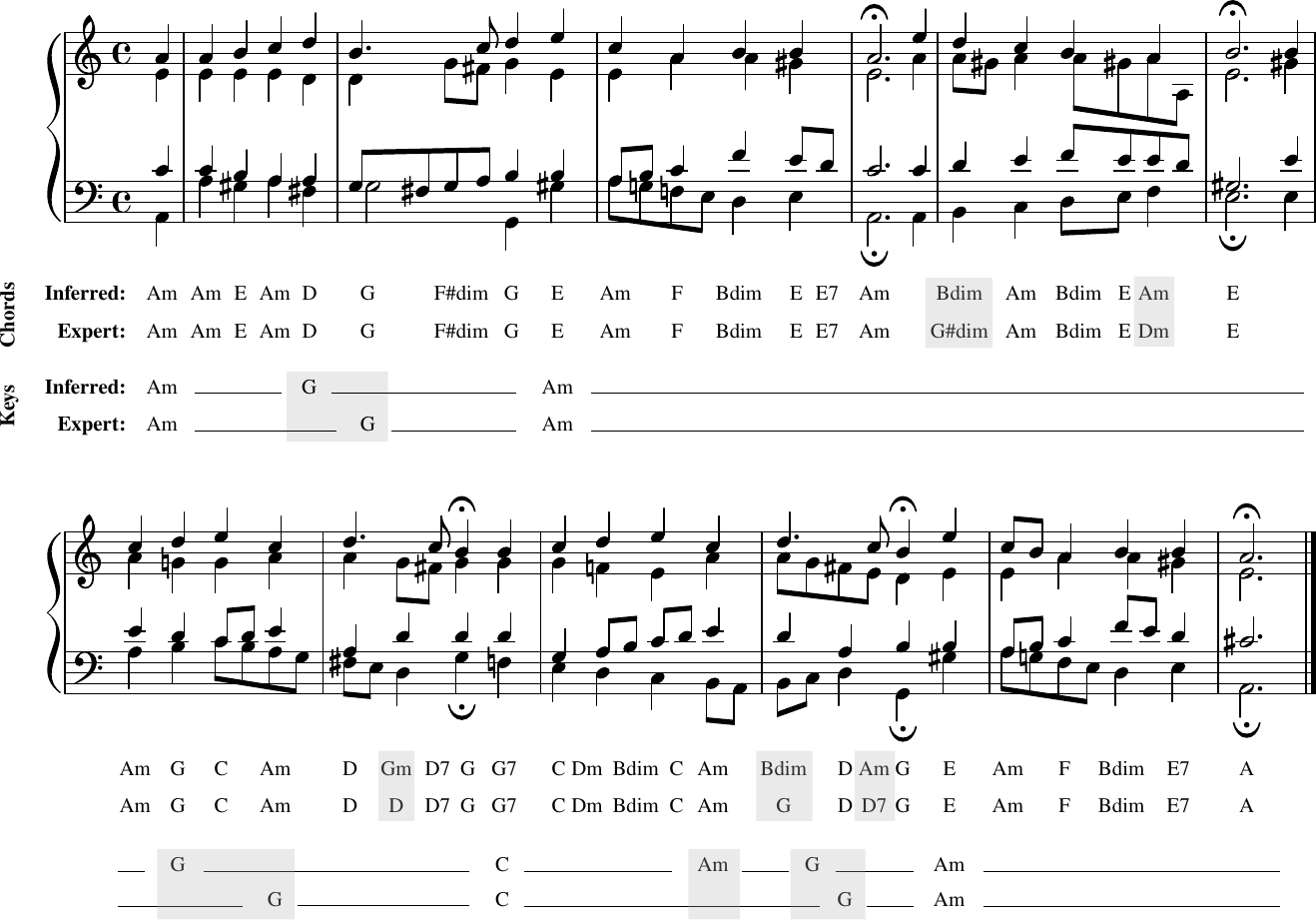}
\end{center}
\caption{Riemenschneider No.\ 88, \textit{Helft mir Gottes g\"ute preisen,} with the chords and keys inferred by our calculation along with chords and keys as assigned by a human analyst.  Points of disagreement are highlighted with gray boxes. }
\label{fig:rs88}
\end{figure*}

The third category of chord, represented by the bottom panel in Fig.~\ref{fig:pnote}, has high probabilities for the anchor pitch, the minor third, and the diminished fifth, suggesting a diminished triad, but also for the minor sixth (eight semitones above the anchor), suggesting a dominant seventh.\footnote{Note that the anchor in this case---the most probable note---is, from a conventional perspective, the \emph{third} of the dominant seventh.}  Thus the third category captures both diminished and dominant seventh chords, a sensible marriage since diminished chords often play a role functionally similar to dominant sevenths, particularly in a major mode.  Note that this category can also represent a pure major triad, with notes 0, 3, and 8 semitones above the anchor---a dominant triad without a seventh.

Once the probabilities are determined, we calculate the chords themselves.  Figure~\ref{fig:rs88} shows a typical example, for Riemenschneider No.~88, along with a human analysis produced by one of us (DT).  Overall, our algorithm does a good but not perfect job of identifying chords.  Out of 45 chords, the calculation gets 40 correct and 5 wrong (highlighted in the figure).  Often the errors are understandable.  For instance, identifying a B\,dim at the start of measure~5 is certainly reasonable, though most human annotators would consider the G\sh\ to represent a chord change.  (In part this is because the progression A minor-B diminished-A minor is rare.)  Less obvious is the A~minor instead of D~minor on the final beat of the same bar: here the algorithm takes the presence of the A-E fifth to suggest an A minor chord, whereas the human considers the E a nonharmonic suspension.

\begin{figure}
\begin{center}
\includegraphics[width=\columnwidth]{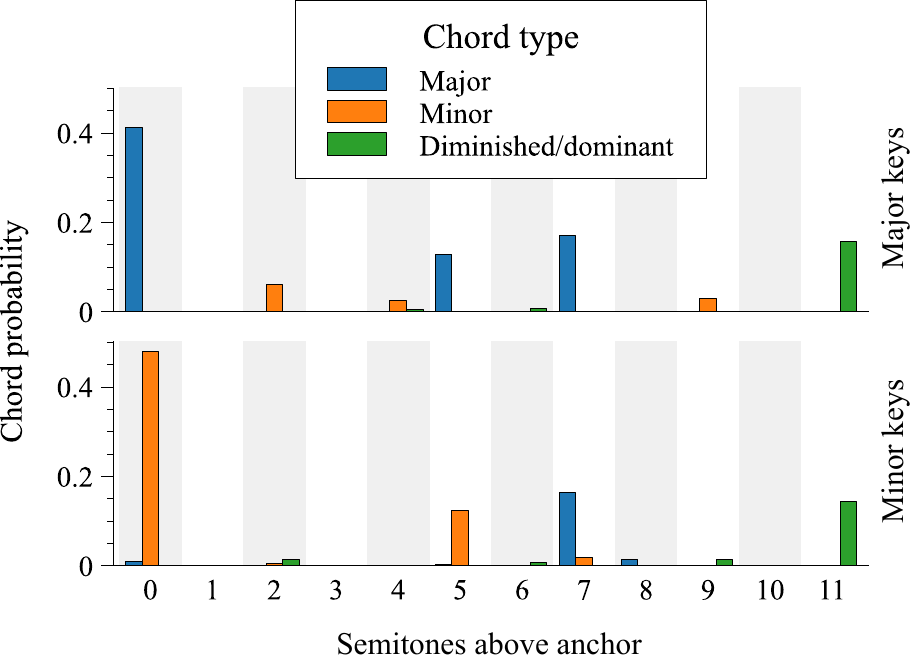}
\end{center}
\caption{Chord probabilities that best capture common patterns in the corpus, for each of the two key types.  Each panel shows the probability of occurrence of each of the three chord types in each of the 12 possible positions relative to the anchor pitch.  The key types closely match traditional major and minor keys.  Vertical gray bars are included merely as a guide to the eye.}
\label{fig:pchord}
\end{figure}

Turning to the keys, our calculation looks for keys of two types, characterized by the probability of occurrence of chords of each of the three types in each possible transposition relative to the anchor pitch.  Again the calculation finds the probabilities first.  The values of the probabilities are shown in Fig.~\ref{fig:pchord} and correspond closely to our expectations for major and minor keys.  In the first key type, shown in the upper panel of the figure and corresponding to major keys, major triads occur with high probability at zero, five, and seven semitones above the anchor and minor triads---with somewhat lower probabilities---two, four, and nine semitones above the anchor.  Diminished/dominant seventh chords are almost always 11~semitones above the anchor, corresponding to a dominant seventh on the fifth or a diminished on the leading tone (which play similar roles).

In the second key type, shown in the lower panel of Fig.~\ref{fig:pchord} and corresponding to minor keys, major triads occur mostly on the fifth scale degree, minor triads on the first and fourth, and diminished on the major seventh again.  All of these probabilities agree well with our expectations.

Over the 50 repetitions of the calculation with different random initial values of the parameters, these are the only probability distributions of chords found by our algorithm.  No other distribution was found even once, indicating with high certainty that this is the best fit to the data.

In passing, we note that although our algorithm operates exclusively in chromatic space, these chord probabilities allow us to infer the presence of familiar scales.  Taking the most probable notes in the most probable chords gives us the diatonic scale above the major-key tonic, (C, D, E, F, G, A, B) in C major, and the harmonic minor scale above the minor-key tonic, (A, B, C, D, E, F, G\sh) in A minor.  This suggests the presence of an additional hierarchical scalar level intermediate between chord and chromatic aggregate.

Once the probabilities are determined, the algorithm makes an estimate of the prevailing key at each step in each work and, like the chords, the results are accurate but not perfect.  Again Fig.~\ref{fig:rs88} gives an example, with disagreements highlighted.  The main error---and one that we see repeatedly in our results---is that our calculation tends to locate key changes slightly earlier than the human analysis does.  For example, the first key change from A~minor to G major is marked on the fourth beat of measure~1 by our algorithm but on the first beat of measure~2 by the human analysis.  The algorithm changes keys at the moment the chords become more probable in the new key than in the old, whereas humans typically default toward remaining in a single key, requiring more definitive confirmation that a new key has arrived (e.g.,~a V-I progression in the new key).  It is debatable whether this is really a disagreement: there are arguably two separate questions here, one about when the key-change process begins, and the other about when it ends; our algorithm answers the first question, whereas human analysts often answer the second.  Furthermore, even expert listeners may differ by a beat or two on exactly when a modulation occurs.

A more significant disagreement is seen at the end of measure~9, where our calculation identifies a modulation to A~minor that the human analysis omits.  This also is a fairly common issue: the computer appears willing to identify brief excursions from the current tonal center as modulations, where the human analysis considers the key to be unchanged.  In this case the algorithm sees A~minor and B\,dim chords as implying a temporary key change to A~minor.  The issue here is that the algorithm has no notion of ``being in one key while briefly using progressions characteristic of another''; it simply calculates the key most likely to produce each sequence of chords.

Once our calculations are complete we can make a quantitative estimate of the accuracy of our method by comparing all chords and keys between the human and computer analyses of all 371 chorales.  For the chords there are a number of ways to perform this comparison, as we discuss in Section~\ref{sec:improvements}, but the simplest is to take each human-annotated chord and translate it into one of the three categories used by the computer---major, minor, and diminished/dominant 7th---then compare this to the category found at the same point by our algorithm.  When we do this, we find agreement in 87.9\% of cases, averaged over all 371 chorales.  Comparing keys is more straightforward, since the computer's two categories clearly correspond to standard major and minor, so no translation between human and machine language is necessary.  We do allow some flexibility about the timing of key changes: we do not penalize cases in which the computer analysis finds the keys correctly but there is disagreement about the timing of the change by up to two steps (as in measures 1 and~2 of Fig.~\ref{fig:rs88}).  With this provision, we find agreement of keys 84.3\% of the time.  These numbers are comparable with the accuracy of previous automated approaches for finding chords and keys, and in the vicinity of what we estimate, intuitively, to be the level of agreement among human annotators.  Exact comparisons with previous work are difficult because other approaches have typically focused on determining either the chords or the keys, but not both, and have used a variety of data sets varying in size, content, and style.  Burgoyne and Saul~\cite{BURGOYNE2005} report 83\% accuracy in determining chords from pitch-class profiles, a little worse than our result; Sheh and Ellis~\cite{SHEH2003} report 75\% on a similar task.  N\'apoles L\'opez~\etal~\cite{NAPOLESLOPEZ2019} summarize a number of previous attempts at key identification, some of which achieve somewhat higher accuracy than ours, though without inferring chords, and in some cases using smaller or simpler data sets or hand-tuned parameter values.

\begin{figure}
\begin{center}
\includegraphics[width=7cm]{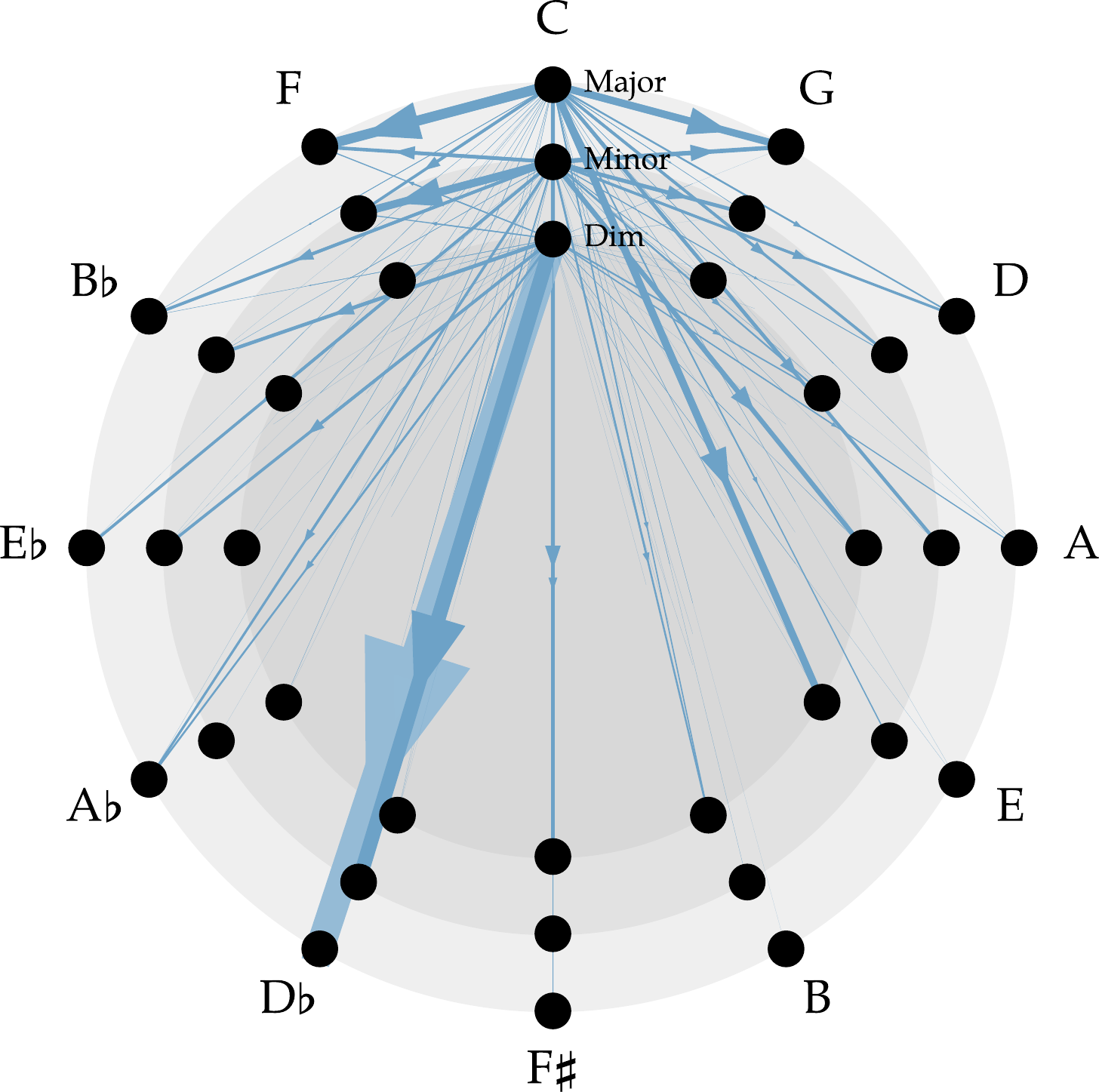}
\end{center}
\caption{Transition rates between chords are represented by the thickness of the arrows in this figure.  The chords are represented by the dots, which are arranged in three circle-of-fifths rings, one for each of the three chord types, major, minor, and diminished/dominant seventh.  All transitions are transposed so that they start from a C anchor, i.e.,~they are represented as being from C~major, C~minor, or C~dim/A\fl7 to another chord.  Thus, for instance, the most common transition is from C~dim/A\fl7 to D\fl\ major, a V7--I cadence.}
\label{fig:chordcircle}
\end{figure}

\subsection{Probabilities for transitions between chords and keys}
Each of our two hidden Markov models also contains probability parameters that represent the likelihood of transitions, between chords on the one hand and keys on the other.  Figure~\ref{fig:chordcircle} depicts the chord-transition probabilities.  The chords, represented by the dots in the figure, are arranged around a circle of fifths and divided into three concentric rings, corresponding to the three types, major triads in the outermost ring, then minor triads and diminished/dominant seventh chords.  Transitions, represented by the arrows, are transposed so as to begin at a chord with a C~anchor.  Thus a transition up a fifth is represented by an arrow from C to~G; down a fifth is represented by an arrow from C to~F.  The varying probabilities themselves are represented by the thickness of the arrows.

Once again, the results found by the algorithm correspond closely to accepted understanding of chord movements, with strong transitions of major chords up and down a fifth (C~major \goesto\ F~major and C~major \goesto\ G~major) and minor chords down a fifth (C~minor \goesto\ F~minor) and a third (C~minor \goesto\ A\,dim, the latter potentially standing in for~F7).  The most frequent diminished-chord transitions are C\,dim \goesto\ D\fl~major and C\,dim \goesto\ D\fl~minor, with C\,dim here potentially standing for A\fl7, so that these correspond to V7 \goesto ~I, V7 \goesto ~i, \s{vii}{}{} \goesto ~I, or \s{vii}{}{} \goesto ~i.

\begin{figure}
\begin{center}
\includegraphics[width=7cm]{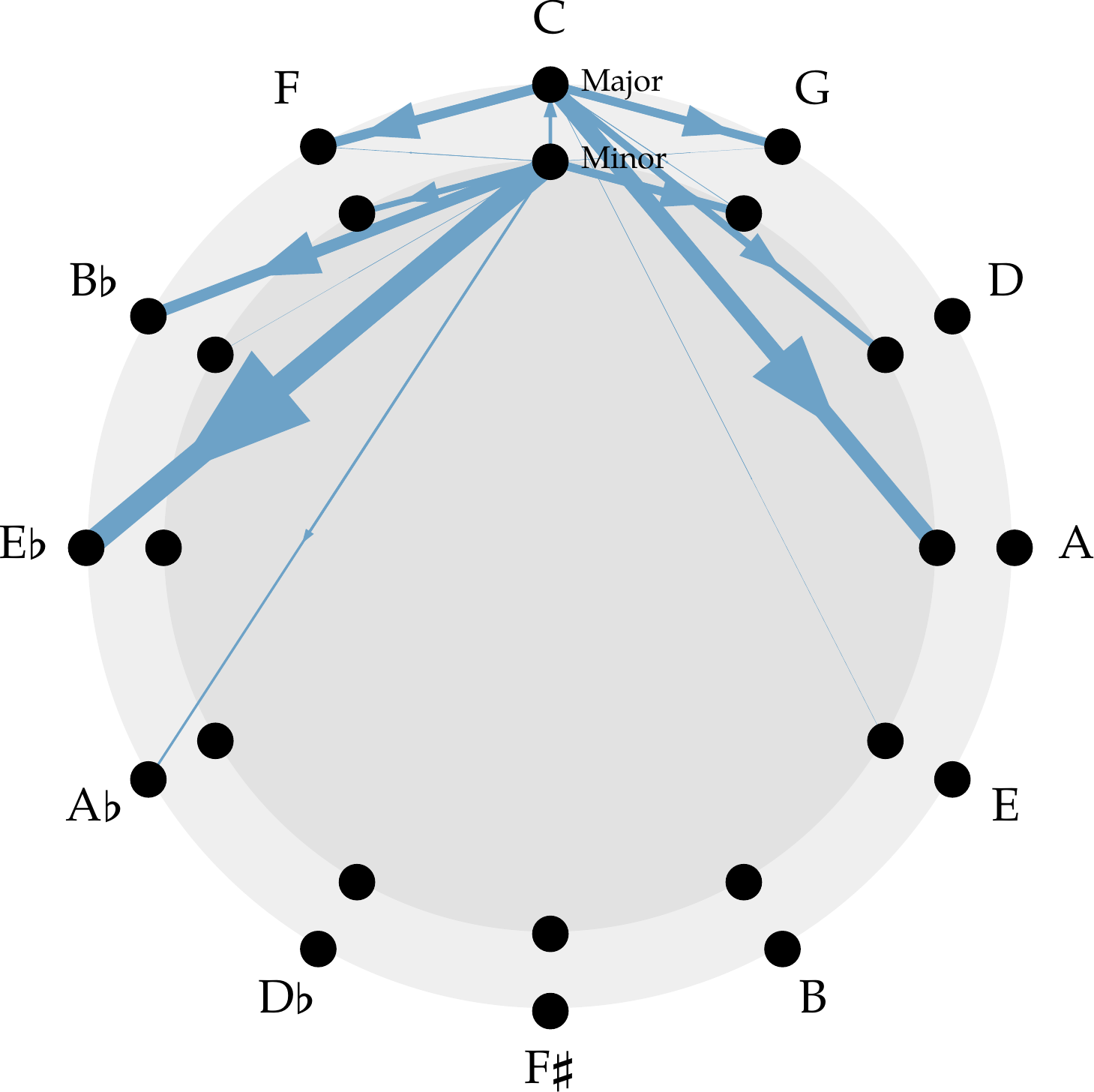}
\end{center}
\caption{Transition rates between keys are represented by the thickness of the arrows in this figure.  The dots represent the keys, arranged in two circle-of-fifths rings for major and minor, and key changes are transposed so that they start from a C~major or C~minor.  Thus the most frequent key changes are C~minor \goesto\ E\fl~major and C~major \goesto\ A~minor.}
\label{fig:keycircle}
\end{figure}

The corresponding results for transitions between keys are shown in Fig.~\ref{fig:keycircle} and again coincide with intuition, the most frequent changes being between the relative major and minor (C~minor \goesto\ E\fl~major and C~major \goesto\ A~minor), one step around the circle of fifths (C \goesto\ G and C \goesto\ F), and two steps (C~major \goesto\ D~minor and C~minor \goesto\ B\fl~major).

While these probabilities are broadly consistent with standard music theory, the correspondence is somewhat indirect.  This is because the probabilities are defined in terms of the movement of the anchor---a C~major chord goes to F~major for instance, or the key of C~major goes to D~minor---but not relative to any key center.  By contrast, music theoretical concepts are typically defined in a key-relative fashion: the tendency of the dominant chord to progress to a tonic, or for the dominant key to modulate to the relative minor.  (In the case of chord progressions the relevant key is the local key while in the case of modulations it is the global tonic of the entire piece.)  Our model's probabilities thus average over syntactically distinct transitions: the probability of the chord progression C major \goesto\ F major is an average of I-IV in C, V-I in F, and so on.

\section{Machine and human analyses}
\label{sec:improvements}
In Section~\ref{sec:results} we evaluated our model's performance by translating human annotations into the language of the hidden Markov model.  In this section we do the reverse, translating the model output into standard harmonic labels, with the goal of both examining the algorithm's performance in greater detail and also of producing an annotated corpus potentially useful to other researchers.

There is some disagreement about the optimal terminology for harmonic analysis.  Generally speaking, music theorists have taken one of three approaches: functional theories use flexible chord categories that do not completely determine a chord's note content (e.g.,~``dominant''); scale-degree theories use Roman numerals that presuppose tertian (``stack of thirds'') structure (e.g.,~``V7''); and figured-bass theories simply list intervals above a bass scale degree (e.g.,~``a sixth, fourth, and third above the supertonic'').  These competing approaches complicate the attempt to find a ground truth against which we can compare our model, as the choice of terminology is itself somewhat controversial.

The target language we use here combines Roman numerals indicating a chord's pitch-class content relative to the key with figured-bass symbols that indicate both the bass note and the presence or absence of a seventh.  Table~\ref{tab:symbols} summarizes the labeling system, which is described more fully in~\cite{TYMOCZKO2019}.  These categories are sharp, imposing strict limits on which notes can be considered harmonic at any time: in C major the symbol~\s{V}{6}{5} corresponds to a harmony containing the notes \{B, D, F, G\} with B in the bass.  Notes may be omitted but no other notes can be included within the harmony.  (To be sure, the style permits ``nonharmonic'' tones---suspensions, neighbors, passing tones, pedals, and anticipations---but these are typically understood as extraneous to the harmony, and subject to additional contrapuntal laws.)  The sharpness of these labels contrasts with the more flexible categories used in jazz and other styles, which permit the addition of non-notated harmonic tones such as ninths, elevenths, and thirteenths.

\begin{table}
\begin{center}
\renewcommand{\arraystretch}{1.2}
\begin{tabular}{lcl}
Triad quality & Chord symbol & Seventh (if present) \\
\hline
Diminished & \s{v}{\dm}{} & Diminished \\
Minor & v & Minor \\
Major & V & Minor \\
Augmented & \s{V}{+}{} & Minor \\
Diminished & \s{v}{\hdm}{} & Minor* \\
\multirow{2}{*}{Any} & \s{v}{\dm\textrm{maj}}, \s{v}{\textrm{maj}}{} & \multirow{2}{*}{Major*} \\
 & \s{V}{\textrm{maj}}{}, \s{V}{\textrm{+maj}}{} \\
\end{tabular}
\\
*Presence of seventh implied by symbol
\null\vspace{3ex}
\renewcommand{\arraystretch}{1.5}
\begin{tabular}{clc}
Triad symbol & Bass note & Seventh symbol \\
\hline
$_3^5$    & Root    & $^7$ \\
$^6$ & Third   & $_5^6$  \\
$_4^6$    & Fifth &   $_3^4$ \\
--   & Seventh & $^2$ \\
\end{tabular}
\end{center}
\caption{Harmonic symbols used in the ground-truth analysis of the Bach chorales.}
\label{tab:symbols}
\end{table}

Our two-level hidden Markov model approach labels both chords and keys with ordered pairs $(x, y)$, where $x$ represents the anchor pitch and $y$ is a probability distribution denoting a class of chords or keys.  For keys, the interpretation is straightforward: the anchor is the tonic note and the two probability distributions of Fig.~\ref{fig:pchord} give the probabilities of occurrence of various chords in major and minor keys.  As mentioned, this approach matches our ground-truth keys 84.3\% of the time.

The interpretation of chords is less obvious.  Each probability distribution in Fig.~\ref{fig:pnote} encompasses a number of different chords, so there is no straightforward translation from distributions to chord symbols.  An effective strategy is to use the score to disambiguate the analysis, inferring the harmonic symbol that best reflects the sounding notes at each step.  Essentially the hidden Markov model constrains the range of possibilities to the point where simple rules can fill in the remaining details.

In this section we describe three different translations of the hidden Markov model output into chord symbols, ordered by their degree of reliance on the score.  For the interested reader, the results are available for download in the RomanText format, which can be read by software packages such as music21~\cite{TYMOCZKO2019, CUTHBERT2010}.

\paragraph*{Method 1.}  Our first translation assigns the first probability distribution in Fig.~\ref{fig:pnote} to the major triad and the second to the minor triad (ignoring the small probability of a sixth or seventh above the bass).  The third category, however, can correspond to five different chords: diminished triads, diminished sevenths, and half-diminished sevenths with root given by the anchor pitch and dominant sevenths or major triads with root a major third below the anchor.\footnote{In a major key these chords are fairly similar and can often be aggregated without much loss of information.  In a minor key, however, they can be quite different: in A~minor, the diminished triad B-D-F is typically a predominant progressing to~E7, whereas the dominant seventh B-D-F-G is an applied dominant typically progressing to~C.}  We resolve this ambiguity by taking as harmonic all those notes present in the score that are either 0, 3, 6, or 8 semitones above the anchor.  This allows for diminished triads, major triads, and dominant sevenths.  Diminished and half-diminished sevenths are identified by first identifying a diminished triad and then the appropriate seventh from the score.  We identify bass notes as the lowest chord tone in the sounding harmony.  The resulting analyses are available in the file ``TN2022raw.zip.''

\begin{table}
\begin{center}
\renewcommand{\arraystretch}{1.4}
\begin{tabular}{cllcl}
& Human              & Computer & Count & Issue \\
\hline
\begin{rotate}{90}
\hbox{\hspace{-55pt}Major key}
\end{rotate}
& \s{V}{}{}       & (0, 0)   & 271   & suspension \\
& \s{ii}{6}{5}    & (5, 0)   & 89    & sixth/seventh \\
& \s{ii}{7}{}     & (5, 0)   & 58    & sixth/seventh \\
& \s{V}{7}{}      & (7, 0)   & 47    & missing seventh \\
& \s{vii}{\dm6}{} & (5, 0)   & 34    & suspension \\
& \s{V}{6}{5}     & (5, 0)   & 28    & suspension \\
& \s{vi}{7}{}  & (0, 0)   & 24    & sixth/seventh \\
& \s{I}{}{}       & (5, 0)   & 23    & suspension \\
\hline
\begin{rotate}{90}
\hbox{\hspace{-60pt}Minor key}
\end{rotate}
& \s{ii}{\hdm6}{~~5} & (5, 1) & 240 & sixth/seventh \\
& \s{V}{}{}        & (0, 1) & 143 & suspension \\
& \s{ii}{\hdm2}{}  & (5, 1) & 44  & sixth/seventh \\
& \s{iv}{6}{}      & (0, 1) & 39  & suspension \\
& \s{ii}{\hdm7}{}  & (5, 1) & 33  & sixth/seventh \\
& \s{V}{6}{}       & (0, 1) & 31  & suspension \\
& \s{vii}{\dm6}{}  & (5, 1) & 26  & suspension \\
& \s{V}{7}{}       & (7, 0) & 26  & missing seventh \\
& \s{V}{6}{}       & (11, 2) & 23 & seventh error \\
\end{tabular}
\end{center}
\caption{The most common mismatches between the hidden Markov model analysis and the human ground truth.  The pair $(x, y)$ represents anchor~$x$ and chord category~$y$.}
\label{tab:mistakes}
\end{table}

\begin{figure*}
\begin{flushleft}
\hspace{3.2cm}\includegraphics[width=10.2cm]{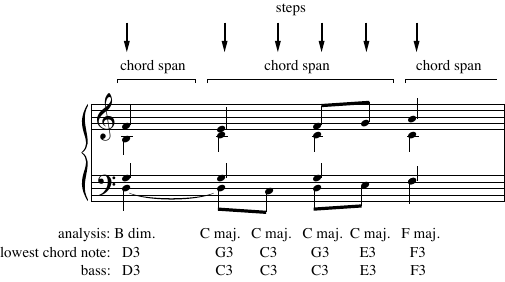}
\end{flushleft}
\caption{A step is a single instant of musical time; any change in any voice creates a new step.  A chord span is a contiguous sequence of steps that have been assigned the same chord symbol.  This figure shows how our translation scheme exploits this distinction to assign bass notes to a passage.}
\label{fig:DT3}
\end{figure*}

This simple translation matches our ground-truth chords 79\% of the time, conditional on getting the key correct.  The lower success rate reflects the fact that this comparison is more demanding than the one we performed in Section~\ref{sec:results}, where we asked only whether the hidden Markov model has assigned each chord to the correct one of the three categories.  Here a correct result requires identifying key, Roman numeral, bass note, and the presence of a seventh.  Table~\ref{tab:mistakes} shows the most common mistakes made by the algorithm, relative to the human ground truth.  In this table ``suspension'' refers to the misanalysis of a chord produced by a suspension, ``sixth/seventh'' refers to the analysis of a seventh chord as a triad with added sixth, and ``missing seventh'' refers to the neglect of a harmonic seventh present in the music.

\paragraph*{Method 2.}  We now consider several ways to improve the translation, by developing better rules that more accurately reflect human judgment.  As we will see, many of these rules reflect notational conventions rather than substantive syntactical questions.

\paragraph*{Keys.}  One issue with the key identification is that the hidden Markov model tends to modulate more readily than a human analyst would.  We can improve the accuracy somewhat, relative to the human ground truth, by eliminating very small key regions that are sandwiched between two identical keys, such as a stretch of C major followed by a two-chord stretch of D minor, followed by a stretch of C major again.  This modestly improves the fit between the hidden Markov model and our human corpus, from 84.3 to 85.6\%.  It is unclear how the results might be further improved: unlike chord identification, key identification does not seem to be amendable to simple algorithmic fixes. 

\paragraph*{Bass notes.} In Bach's chorales the bass part normally articulates the true bass note even if it briefly crosses above the tenor.  This is because it was often doubled in a lower octave by the organ.  Thus rather than taking the lowest sounding note to be the bass, we should consider the bass voice to be the carrier of the bass note at all times.  Like the other musical voices, however, the bass sometimes sounds nonharmonic tones not belonging to the chord.  To deal with this issue, we distinguish the \emph{step} or instant of musical time from the \emph{chord span}, a maximal contiguous span of steps assigned identical analytical symbols, with the understanding that the true bass note may not be present in every step in a span.  We can define the initial bass note in a given chord span as the first harmonic note in the bass part.  This remains in force until it is displaced by another harmonic note in the bass part within the same span (Fig.~\ref{fig:DT3}).

\paragraph*{Chord labels.} We have seen that the third chord category in Fig.~\ref{fig:pnote} is ambiguous, encompassing a diminished triad, diminished or half-diminished seventh, dominant triad, or dominant seventh a third below.  Close inspection shows that the other two categories are also somewhat ambiguous.  The first, shown at the top of Fig.~\ref{fig:pnote}, has 0, 4, and 7 semitones as its most probable intervals above the anchor, with smaller probabilities for 9 and 11 semitones.  This means that it could potentially represent \{C, E, G\}, \{C, E, G, B\}, \{C, E, G, A\}, or \{C, E, A\}.  Similarly, the second category could represent the minor triad \{C, E\fl, G\}, the minor seventh \{C, E\fl, G, B\fl\}, or the half-diminished seventh \{C, E\fl, G, A\}.  Thus, the same set of notes can be analyzed in multiple ways: \{A, C, E, G\} can be either category 1 or 2 (corresponding to either C major with an added sixth or A minor with an added seventh) and \{A, C, E\fl, G\} can be analyzed as either category 2 or 3 (corresponding to C minor with an added sixth or A diminished with an added seventh).

As we discuss in the next section, this sort of multivalence is characteristic of some harmonic theories.  Here, however, our goal is to match a particular style of analysis in which chord symbols set a strict limit on the allowed harmonic scale degrees.  To do this, we start by distinguishing the sounding notes occurring within a particular step from the \emph{persistent notes}, those sounding in every step of the chord span.  We then develop rules that determine whether a given note is harmonic or not, depending on whether it is found in the persistent notes. The best match to the human analysis requires a slightly different strategy for each of our three chord categories.  

For the first (the ``major chord'') we consider the major sixth or major seventh to be harmonic only if they belong to the persistent notes, but the minor seventh to be harmonic if it belongs to the sounding notes.  This is because the major sixth and major seventh often appear as unlabeled passing tones, while the minor seventh signals a dominant seventh, which is very likely to be labeled in a human analysis.  The final possibility, the major triad with a minor sixth, is very rare in Bach's chorales.

For the second category (``minor chord'') we consider the sixth to be harmonic only if it belongs to the persistent notes, but the minor seventh to be harmonic if it belongs to the sounding notes.  (The major seventh over a minor triad is again very rare.)  For the third category we consider the minor sixth, diminished seventh, and minor seventh to be harmonic if they occur in the sounding notes.  

These rules reflect both notational conventions, such as when to label a seventh, and substantive beliefs about musical structure, e.g.,~that a major chord with a minor seventh almost always represents a dominant seventh chord.  Requiring that a note belong to the persistent notes sets a high standard for harmonicity: such notes only count as chord tones if they last for an entire chord span, in which case they are very unlikely to be nonharmonic.

\paragraph*{Cadential formulas.} A final issue involves idiomatic cadential suspensions that human beings would normally consider nonharmonic, such as (G3, C4, D4) \goesto\ (G3, B3, D4).  Our algorithm often analyzes this progression as two different chords, a C major (missing the E and with a D in the persistent notes) followed by G major.  We can eliminate most of these errors by hand-coding a small set of standard suspension patterns: for instance, if the persistent notes in a category-1 chord include the major second above the anchor (e.g.,\{C, D, G\}), and if the music progresses to another category-1 chord up by fifth (e.g., \{B, D, G\}), then we eliminate the initial chord, extending the second chord backward to cover its predecessor.  These rules are summarized in Table~\ref{tab:suspensions}.  Here it is worth noting that four-chord versions of our model typically settle on suspension formations as a fourth probability distribution, underscoring their importance to the style.

\begin{table}
\begin{center}
\begin{tabular}{cccc}
Semitones  & Orig. anchor  &            & New \\
above anchor & progression & Anchor delta & chord type \\
\hline
0, 2, 7, 9 & 2           & 2          & Major \\
0, 2, 7, 9 & 6           & 6          & Diminished \\
0, 2, 7    & 6           & 2          & Major \\
0, 2, 7    & 7           & 7          & Major \\
0, 5, 7    & 7           & 7          & Major \\
0, 5, 7    & 11          & 11         & Diminished \\
\end{tabular}
\end{center}
\caption{Hard-coded suspension formulas.  The first two columns represent the inputs to the algorithm: ``semitones above anchor'' refers to the notes found above the assigned anchor (which falls at~0 in each case); ``original anchor progression'' is the change of anchor to the next sonority.  The next two columns represent the effects of the algorithm: ``anchor delta'' is the transformation applied to the original anchor and ``new chord type'' is the newly assigned chord type.  Thus the third line says that if the algorithm finds 0, 2, 7 with an anchor progression of 6, then this chord is described as a major chord (category 1) on 2.}
\label{tab:suspensions}
\end{table}

Thus improved, the translation---our second of three---now agrees with the human annotations on about 87\% of the chordal assignments, conditional on agreeing about the key.  (The annotations are available in the file ``TN2022.zip.'')  We have obtained this agreement at the cost of using a modest amount of knowledge about the underlying musical style: without this it would be difficult to translate the three-category output of the hidden Markov model into the much richer symbol system of standard harmonic analysis.  Importantly, this knowledge is embodied by general rules that are independent of local musical context: relative to key, the same set of notes is always treated the same way. 

\paragraph*{Method 3.}  Our third translation improves accuracy still further by allowing for ``nonharmonic consonances''---triads and seventh chords produced by nonharmonic tones and not part of the harmonic narrative.  (Gottfried Weber called these ``apparent chords''~\cite[pp.~658--678]{WEBER1846}.)  These nonharmonic consonances are an important source of the holism mentioned in the introduction: choosing between the analyses in Figs.~\ref{fig:ambiguity} and~\ref{fig:huron} requires a sense of which harmonic progressions are likely to occur, expectations that must be in place prior to analyzing a given passage.  Our third method thus introduces a degree of context-dependency into the analyses.

We can simulate background knowledge using the results of our hidden Markov model analysis.  First we count every two-chord progression in the dataset annotated using our previous method and use these counts to estimate the frequency of each progression.  Second, we encode a basic set of rules about the nonharmonic tones permissible in Bach's style---the ``contrapuntal grammar'' that was largely inherited from the Renaissance.  (Thus we stipulate, for example, that passing and neighboring tones are out-of-chord notes that are preceded and followed by chord-tones to which they are connected by step.) These rules give us a set of contrapuntally permissible analyses for every step.  Third, whenever we find two eighth-note harmonies occurring within a single quarter-note beat, we ask whether either of those harmonies could be analyzed as nonharmonic.  This gives us a range of possible chord progressions, from which we choose the one whose chord progressions have the highest average frequency in our count of two-chord progressions.  The result can be thought of as the ``most likely analysis,'' given harmonic expectations based on the raw hidden Markov model output and the additional rules described here.

\begin{figure*}
\begin{center}
\includegraphics[width=11cm]{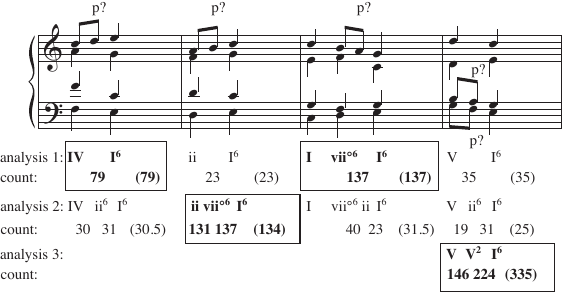}
\end{center}
\caption{Four one-measure passages analyzed in multiple ways.  In the first analysis, the tones marked ``p?''\ are treated as nonharmonic passing tones.  In the second, they are treated as harmonic tones.  The third treats the tenor note as passing and the bass note as harmonic.  Underneath each progression we show the number of times it occurs in the corpus.  The numbers in parentheses are averages.  Our algorithm chooses the progression with the highest average, enclosed in boxes.}
\label{fig:DT5}
\end{figure*}

Figure~\ref{fig:DT5} shows how the algorithm treats four superficially similar passages.  Notes marked ``p?''\ are flagged as potentially nonharmonic and we consider one reading in which the note is a passing tone and another in which it is harmonic.  The numbers underneath each two-chord progression show the number of occurrences of this progression in our previous analysis.  The algorithm chooses the reading with the highest average frequency, shown in bold and surrounded by a box.  This sometimes requires treating superficially similar passages in different ways.  Our third approach achieves this context-dependency in a rule-governed way, relying on first-pass statistics to generate a most-probable interpretation from among a set of contrapuntally permissible alternatives.  By contrast, human beings make context-dependent judgments whose origin and justification can be hard to understand.

This pruning of the analysis improves accuracy to the point where we match the human chord symbols 90\% of the time, assuming agreement about the key.  The resulting data set, which is accurate enough for a range of music-theoretical purposes, is available as ``TNpruned.zip.''  The high degree of agreement with the human analysis demonstrates that analytical competence does not require much training or experience.  Instead it can be learned from a suitable corpus of music, along with a modest amount of style-specific knowledge.  The subtle analytical judgments that go into Fig.~\ref{fig:DT5} can be simulated with relatively presupposition-free methods, making almost no assumptions about the specific harmonic syntax of the music.

\section{Tests of Western harmonic theory}
\label{sec:discussion}
The most obvious use for these datasets is as a neutral testing-ground for theories of Western harmonic structure.  The biggest question here is whether traditional music has a chord-to-chord harmonic structure akin to that of a first-order Markov model, articulated by rules of the form ``V chords generally go to I chords but not IV chords.''  Many theorists believed such structure exists, including Rameau, Weber, Riemann, Piston, Kostka, Payne, and other contemporary pedagogues.  A more recent alternative is that harmony is characterized by a hierarchical structure rooted in melodic laws---a view originating with Heinrich Schenker~\cite{SCHENKER1997} and endorsed by post-Schenkerians such as Lerdahl and Jackendoff~\cite{JACKENDOFF1983}.  In recent years, this alternative perspective has been incorporated into introductory textbooks~\cite{LAITZ2008, CLENDINNING2016}.

\begin{figure}
\begin{center}
\includegraphics[width=4.5cm]{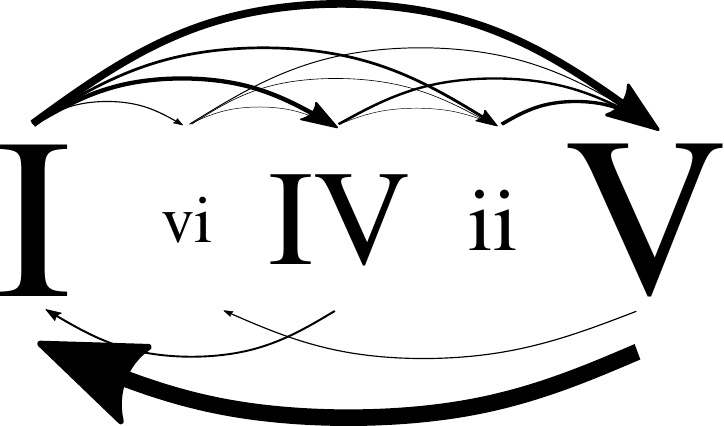}
\end{center}
\caption{The most common chords and progressions in the major-mode passages of the Bach chorales.  Size of chord symbols indicates their frequency of occurrence; width of arrows indicates frequency of transitions.}
\label{fig:chordprogs}
\end{figure}

Our data provide strong evidence for the existence of chord-to-chord syntax.  The most popular chords and transitions found in the major-mode passages of the chorales can be visualized as shown in Fig.~\ref{fig:chordprogs}.  In this figure, symbol size corresponds to frequency while arrow size indicates the relative frequency of the transitions.\footnote{The figure shows Roman numerals accounting for at least 5\% of all chords in the chorales' major-key passages (disregarding inversion), along with all transitions accounting for at least 0.75\% of all transitions.} This representation is reminiscent of the prescriptive maps found throughout the theoretical and pedagogical literature~\cite{PISTON1941,KOSTKA2003, TYMOCZKO2011, TAOM}.  While the presence of this chord-to-chord structure does rule out additional hierarchical organization, it does suggest that any such organization must either produce or complement the music's nonhierarchical organization.  This is significant insofar as Schenkerian theorists have sometimes rejected~\cite{SCHENKER1997} or neglected~\cite{LERDAHL2001} this sort of chord-to-chord structure.

\begin{figure}
\begin{center}
\includegraphics[width=\columnwidth]{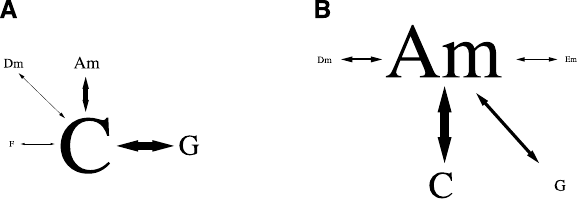}
\end{center}
\caption{The most common modulations in the chorales, for (A)~major keys normalized to C major and  (B)~minor keys normalized to A~minor.  Size of chord symbols indicates frequency of occurrence.  Width of arrows indicates frequency of modulations.}
\label{fig:keyprogs}
\end{figure}

Figure~\ref{fig:keyprogs} provides the same information for modulations, transposing all major-key pieces to C major and all minor-key pieces to A minor.\footnote{In making this graph we selected only those chorales that start and end in the same key, transposing them to so that the starting key was either C major or A minor.} Here the most common modulations are oscillatory: keys tend to move from the tonic to a subsidiary key and back, rather than on longer journeys through multiple subsidiary keys.  This is likely a feature of the chorales in particular, and not characteristic of genres such as the classical sonata.  In this respect the modulatory grammar is unlike the chord-to-chord grammar, which describes a much wider range of harmonic practices---indeed, a good portion of eighteenth- and nineteenth-century music.

Another longstanding area of theoretical controversy is the principles governing ``doubling,'' or the duplication of elements in four-part triadic harmony.  Aarden and von Hippel have surveyed the astonishing range of disagreement surrounding this subject~\cite{AARDEN2004}.  Table~\ref{tab:doubling} shows which notes Bach tends to double in major mode, as extracted from our automated analysis.  In all cases we find that the bass is the most frequently doubled, unless it is a leading tone in which case it is not doubled.  Note for instance that the bass of the first-inversion tonic chord is frequently doubled despite longstanding injunctions to the contrary.  For all the controversy surrounding the subject, Bach's practice is fairly clear.

\begin{table}
\begin{center}
\begin{tabular}{lccc}
Chord              & Bass & Third/fourth & Fifth/sixth \\
\hline
\s{I}{}{}       & 92   & 6            & 2 \\
\s{I}{6}{}      & 47   & 29           & 23 \\
\s{I}{6}{4}      & 62   & 21           & 17 \\
\s{ii}{}{}      & 66   & 27           & 7 \\
\s{ii}{6}{}     & 57   & 21           & 22 \\
\s{iii}{}{}     & 62   & 35           & 3 \\
\s{iii}{6}{}    & 81   & 8            & 10 \\
\s{IV}{}{}      & 75   & 19           & 6 \\
\s{IV}{6}{}     & 65   & 18           & 17 \\
\s{V}{}{}       & 91   & 2            & 7 \\
\s{V}{6}{}      & 8    & 44           & 48 \\
\s{vi}{}{}      & 60   & 35           & 5 \\
\s{vi}{6}{}     & 77   & 18           & 5 \\
\s{vii}{\dm}{}  & 11   & 89           & 0 \\
\s{vii}{\dm6}{} & 74   & 25           & 1 \\
\end{tabular}
\end{center}
\caption{The most commonly doubled notes in the chorales, for chords occurring more than 40 times (0.3\%) in the chorales.}
\label{tab:doubling}
\end{table}

There are other potential analytical uses for our data.  Combined with the musical score, our analysis allows one to algorithmically catalog nonharmonic tones (Table~\ref{tab:DT8}), identify the specific voice leadings used in a given harmonic progression, explore the resolution or nonresolution of tendency tones (e.g.,~leading tones and sevenths), detect parallel perfect intervals either on the musical surface or masked by nonharmonic tones, and search for examples of various phenomena of interest (e.g.,~unusual chord progressions).  In this way our data provides a quantitative tool for grounding theoretical assertions that have traditionally been supported intuitively.

Our work also suggests some more general theoretical lessons.  Theories of Western harmony are subject to a tradeoff between the determinacy of chord categories and the complexity of chord grammar.  This issue dates back to Jean-Philippe Rameau (1683--1784), celebrated as ``the Newton of music'' and in many ways the progenitor of contemporary harmonic theory.  Rameau is sometimes credited with being the inventor of ``inversional equivalence,'' the thesis that pitch-class content determines chordal identity.  From this perspective the notes \{C, E, G\} always constitute a C-major chord no matter how they are arranged in register.  When pitch classes are labeled relative to a tonal center, this implies that a chord's scale-degree content determines its harmonic category.  In fact, however, Rameau was not a pure scale-degree theorist, and his chord-categories had a flexible relation to pitch-class content.  For example, his 1722 \textit{Treatise on Harmony} explained the deceptive cadence V7-vi as a descending-fifth progression in which the second chord was represented not by the root, third, and fifth but by root, third, and sixth~\cite[bk.~2, ch.~6]{RAMEAU1971}.\footnote{Rameau later abandoned this analysis of deceptive cadences, although followers such as d’Alembert continued to uphold it~\cite{CHRISTENSEN1993}.  However, Rameau never abandoned the analogous analysis of the predominant chord with added sixth, arguing that \{F, A, C, D\} in C major could have a fundamental bass of either F or D, or indeed both.}  Thus on this view the notes \{A, C, E\} could represent either a chord whose root is A or a chord whose root is C.

\begin{table}
\begin{center}
\begin{tabular}{lc}
Type                          & \% \\
\hline
Passing (ascending)           & 26 \\
Passing (descending)          & 23 \\
Suspension                    & 13 \\
Accented passing (descending) & 7 \\
Neighbor (descending)         & 6 \\
Unidentified                  & 4 \\
Accented passing (ascending)  & 3 \\
Pedal                         & 3 \\
Anticipation                  & 2 \\
\end{tabular}
\end{center}
\caption{Nonharmonic tones in the chorales.}
\label{tab:DT8}
\end{table}

Here Rameau opted for a simpler harmonic grammar at the expense of a more complex categorization scheme: he described both the authentic and deceptive progressions with a single rule, ``G7 chords progress down by a fifth to C chords.''  But this simplification came at the cost of a categorization scheme in which chordal identity could not be identified from note content alone.  Instead, context determined whether the notes \{A, C, E\} represented C major or A minor.  This can be considered a \emph{functional} approach to chord categorization, as a chord's label depends on its local behavior.  Functional approaches can be controversial insofar as they identify chords on the basis of pre-existing theory: \{A, C, E\} might be identified as a kind of C chord when preceded by G7 but a kind of A chord when preceded by E7.  Thus the claim that chords descend by a fifth is not simply a hypothesis supported by data, but a rule that regulates analysis.  

Something similar can be said of Hugo Riemann, one of the most influential nineteenth-century theorists.  Riemann considered the basic chords to be tonic, subdominant, and dominant, or I, IV and V in Roman numeral notation, arguing that tonal phrases typically articulate a series of I-IV-V-I progressions~\cite{BURNHAM1992}.  However these basic chords were subject to category-preserving transformations: the ``parallel'' transform replaces a chordal fifth with the sixth above the root, and the ``leittonwechsel'' replaces a chordal root with the note a semitone below.  This allowed Riemann to describe both G7 \goesto\ C and G7 \goesto\ Am as instances of a single underlying dominant-to-tonic pattern.  In G7 \goesto\ C, the tonic is represented by its primary triad, while in G7 \goesto\ Am it is represented by the primary triad's parallel variant.  Once again, the result is an ambiguity at the level of chord categories: the chord \{A, C, E\} can represent either a tonic (with A replacing the fifth of \{C, E, G\}) or subdominant (with E replacing the root of \{F, A, C\}).  

Our unsupervised hidden Markov model learns a series flexible chord categories that resemble those of Rameau and Riemann.  Like those theorists, it groups the diminished triad \{B, D, F\} with the dominant seventh \{G, B, D, F\} into a single category.  Like Rameau, it analyzes the notes \{D, F, A, C\} sometimes as an F major chord with added sixth and sometimes as a D minor chord with added seventh.  This is not because it considers these chords to ``sound similar,'' nor because it is committed to metaphysical claims about the priority of descending-fifth progressions.  Rather it does so solely for reasons of information-theoretic efficiency, mutually adjusting probability distributions and transition matrices so as to maximize its ability to predict the future.  What is surprising is that this process reproduces results familiar from the history of music theory.  Those claims have sometimes been denigrated as methodologically disreputable~\cite[p.~192]{CHRISTENSEN1993}, but our work suggests they can be given an algorithmic foundation: musical ``function'' can be understood information theoretically, as the maximally parsimonious description of the input data.\footnote{This result is consistent with work of Jacoby, Tishby, and Tymoczko~\cite{JACOBY2015}, who use the “information bottleneck” to cluster Roman numerals into categories, recovering something resembling traditional functional theory.}

Contemporary North American theory tends to prefer more determinate chord labels at the cost of a more complex harmonic grammar: in this approach, \{C, E, G\} is a C-major chord and \{A, C, E\} is an A~minor chord regardless of context.  This in turn requires a more complex set of harmonic rules according to which G7 progresses most often to \{C, E, G\} (the ``authentic'' resolution) and less often to \{A, C, E\} (the ``deceptive'' resolution).  This approach generally allows for a fine-grained description of musical practice, at the cost of being less enlightening about \emph{why} G7 progresses as it does.\footnote{There are some circumstances where contemporary North American theory uses functional rather than scale-degree labels.  The most common involves cadential $_4^6$ chords such as \{G3, C4, E4\} in C major.  Where a pure scale-degree theory would label this chord \s{I}{6}{4}, many contemporary theorists use \s{V}{6}{4} to record the fact that it typically embellishes and precedes a dominant.  Other circumstances include the assimilation of \s{vii}{\dm}{} to \s{V}{}{} (as a ``dominant seventh lacking its root''~\cite{DAHLHAUS1990}), the distinction between ``plagal'' and ``predominant'' functions of \s{IV}{}{}~\cite{SWINDEN2005}, and the correct labeling of chords like \{G3, B3, E4\} in C~major (which can be considered a dominant or a mediant~\cite{ROTHGEB1996}).}  In standard Roman-numeral theory this is just a fact about the grammar, whereas for Riemann and Rameau \{A, C, E\} is portrayed as being intrinsically similar to \{C, E, G\}.  In this way, the different approaches serve somewhat different purposes: the general categories of Rameau and Riemann provide a partial \emph{explanation} for tonal practice, one that was useful for composers working in the style.  The more precise categories of contemporary theory, by contrast, may help scholars articulate a more detailed grammar of historical practice.

This suggests a pluralistic picture in which there is more than one way to describe harmony.  The most specific simply represents harmonies as collections of scale-degrees above a bass.  This is likely how the earliest functional composers thought---musicians like Corelli, Vivaldi, Handel, and J.~S.~Bach, who wrote before the widespread acceptance of inversional equivalence.  More general are the theory-laden harmonic categories of North American theory, Roman numerals presupposing stack-of-thirds harmonic structure.  These facilitate the description of general patterns that can appear over a variety of different bass notes, such as ``V~chords generally progress to I~chords.''  The even more general theories of Rameau and Riemann use still more flexible categories that help explain the principles underlying the grammar.  Our hidden Markov model natively produces something very much like this last approach, though as we have seen it can be translated into Roman numerals.

These different approaches each have their partisans among contemporary theorists: Gjerdingen~\cite{GJERDINGEN2007} and Holtmeier~\cite{HOLTMEIER2007} have advocated for figured-bass representations; Tymoczko~\cite{TYMOCZKO2011, TAOM} for Roman numerals; and Agmon~\cite{AGMON1995} for harmonic functions.  Yet the different analytical languages are arguably optimized for different musical purposes.  Instead of asking which approach is best, it might be better to ask which is better for a particular application.  In particular, it seems likely that more flexible categories are better suited for introductory or creative contexts, where the goal is to apply broad principles in novel contexts, while the more fine-grained language may be better for detailed syntactical investigation.  In other words, introductory textbooks might benefit from the functional approach, while more specialized investigations might need the language of Roman numerals.

\section{Conclusions}
\label{sec:conclusion}

There is at best a tenuous consensus about the basic concepts of standard harmonic theory.  Schenker and many of his successors deny that harmonies obey any local chord-to-chord grammar, proposing that music is instead governed by purely linear laws.  Functional theorists follow Rameau and Riemann in promoting flexible categories that can be represented by many different triads.  Schoenbergians, along with Ian Quinn and Christopher White, challenge the distinction between harmonic and nonharmonic—--the very picture of tertian harmonies decorated by ``merely linear'' tones.  And figured-bass theorists from J. S. Bach to Holtmeier and Gjerdingen have advocated for figured-bass symbols.

Against this background of widespread disagreement, it is remarkable that our largely assumption-free computation manages to reproduce so many familiar features of traditional theory.  Its two key categories are directly interpretable as major and minor.  Its three chordal categories are broadly synonymous with ``major,'' ``minor'' and ``diminished or dominant.''  Yet at a more fine-grained level these categories admit a degree of flexibility that recall the function-theoretical ideas of Riemann and Rameau: the notes \{F, A, C, D\} can be interpreted either as a D~minor chord with added seventh or an F~major chord with added sixth.  We have seen that it is possible to translate these flexible categories into the more precise terminology of North American theory, where added sixths are unknown and there is a sharp distinction between D minor and F major.  Indeed, our automated analyses largely accord with the human ground-truth annotations.  

Our work results in several datasets that can be used for further music-theoretical and musicological inquiry.  Though perhaps less accurate than purely human analytical corpora, they are also less subjective, being largely independent of human biases: thus we can be fairly sure that regularities appearing in these data are not a mere byproduct of analytical preconceptions.  Furthermore, all the assumptions that go into producing the corpora are clearly spelled out, embodied in the computer programs we have used. Though this paper has not focused on detailed corpus study, we have already observed some non-obvious features of musical structure, such as oscillatory rather than progressive modulatory norms (Fig.~\ref{fig:keyprogs}).  One project for future work is to explore these datasets more thoroughly.

Another is to develop methods that deal with a broader range of music.  Bach's chorales are unusual in having a fixed number of voices, avoiding rests and incomplete chords, using an extremely regular harmonic rhythm, and explicitly delineating voices and phrases in their notation.   Classical-style harmonies, by contrast, change irregularly, are often incomplete, and feature a variable number of voices not explicitly embodied in the notation.  It is not obvious that we should be able to analyze this music with largely assumption-free methods.  Instead, it may be necessary to make further assumptions about classical syntax.  Solving this problem, however, would open the door to a new era for music theory and music history, one that exploits big data and large-scale statistical analysis.

\vspace{1ex}
\begin{center}
\rule{6cm}{0.5pt}
\end{center}

\appendix

\begin{figure*}
\begin{center}
\includegraphics[width=\textwidth]{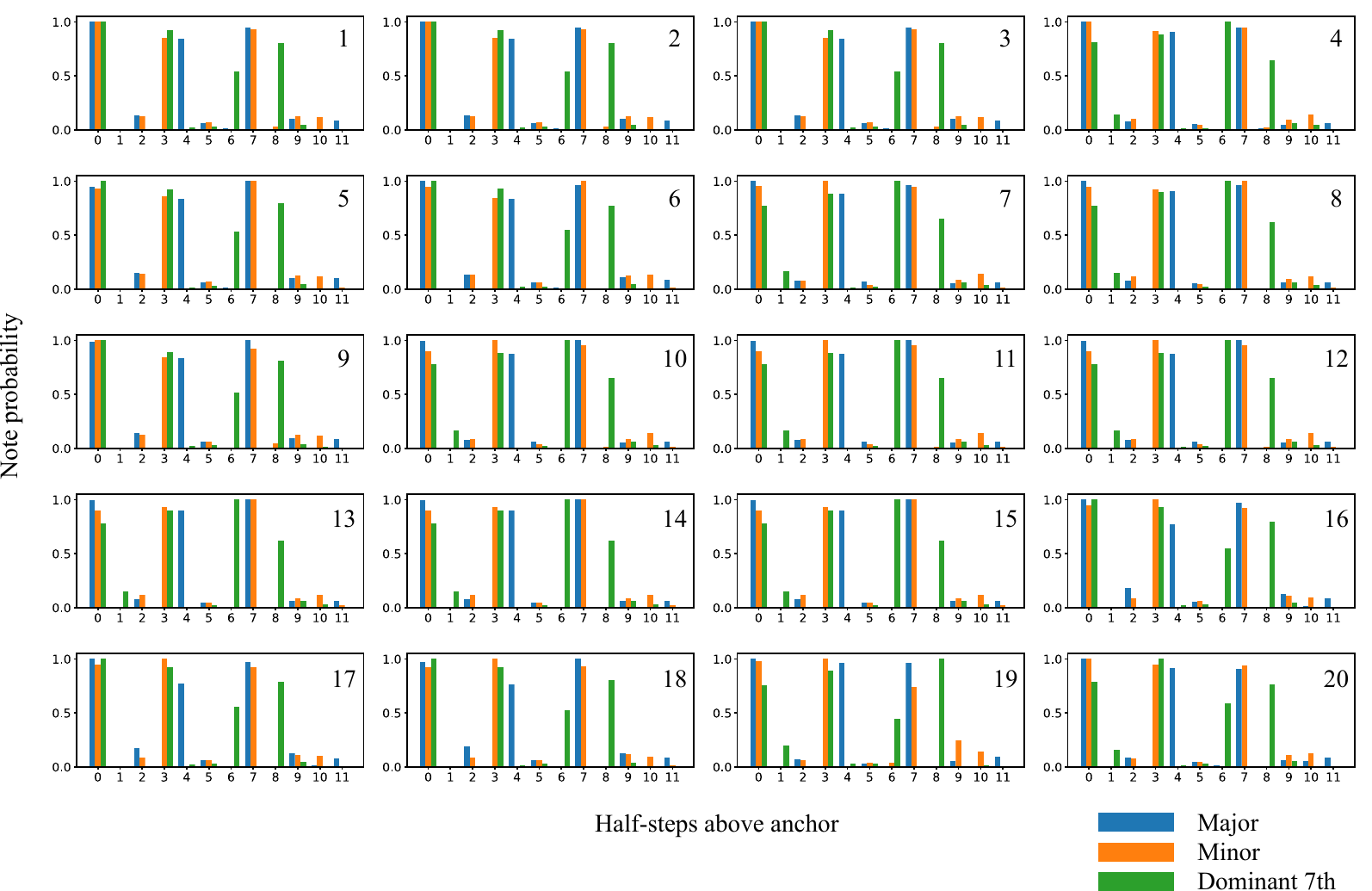}
\end{center}
\caption{Note probabilities for the chord classes found in the 20 runs with highest data likelihoods.  The classes have been manually arranged in order major-minor-dominant seventh in each panel for ease of comparison; in the actual runs the order is random.}
\label{fig:pgrid}
\end{figure*}

\section{Results of individual runs}
\label{sec:runs}
Our calculations consist of two separate parts: the analysis of notes to identify chords and the analysis of the resulting chords to identify keys.  Over multiple repetitions of the calculation the second step---the identification of keys---reliably converges to only a single solution with key classes corresponding to major and minor keys.  No other solution is ever found.  The first step, in which the chords are identified, converges most often to the solution described in Section~\ref{sec:results}, with three chord classes corresponding to major, minor, and diminished/dominant seventh chords.  In 37 of the 50 runs we performed---roughly 75\% of the time---we found these chord classes.  In the remaining 13 runs we found other solutions, typically with two of the conventional chords plus one alternate.  A common alternate, for instance, is the dominant seventh with suspended fourth \{C, F, G, B\fl \}, which occurs in place of the dominant seventh.

These findings, however, somewhat understate our results, in that the alternate solutions, while occurring a moderate fraction of the time, do so with hugely diminished probability.  The probability of these solutions, as quantified by the data likelihood $P(\mathcal{C}|\theta)$ (see Appendix~\ref{sec:likelihood}), is much lower than for the major/minor/dominant solution.  For example, the highest likelihood solution found overall---which has the major/minor/dominant form---has a data log-likelihood of $-118926.5$ (base~$e$).  The best of the alternate solutions has a log-likelihood of $-121413.4$.  This means that even this best alternate is lower in probability by the stupendously large factor $e^{-2486.9} \simeq 10^{1080}$.  Thus the major/minor/dominant classification is very heavily favored in our analysis.

Figure~\ref{fig:pgrid} shows the note probabilities for the chord classes in the top 20 of our calculation runs, those with the highest data likelihoods---each panel is the equivalent for the given run of Fig.~\ref{fig:pnote} in the main text.  Of these twenty plots, every single one corresponds to the major/minor/dominant solution.  In fact, all of the top 31 runs give this solution; one has to go down to the 32nd-best run out of 50 to find one of the alternates.  Again the results speak clearly: the standard major/minor/dominant combination is the best way to describe the structure of these data.

\section{The Baum-Welch algorithm}
\label{sec:bw}
The Baum-Welch algorithm is an expectation-maximization (EM) algorithm for estimating the optimal values of the parameters of a hidden Markov model.  We describe the process here in the language of keys and chords, but the same algorithm applies to any hidden Markov model.

Suppose we have a corpus of $N$ works labeled by $i=1\ldots N$, with the $i$th work having $M_i$ chords $C_i = ( c^i_1, \ldots, c^i_{M_i} )$.  The parameters of the hidden Markov model, which are the same for every work, are the transition probabilities $\beta = \lbrace \beta_{kk'} \rbrace$ and chord probabilities $\nu = \lbrace \nu_{kc} \rbrace$ as defined in Section~\ref{sec:chords}, plus the prior probabilities~$\pi = \lbrace \pi_k \rbrace$ that the first key in a work is~$k$.  (In our calculations we assume $\pi_k$ to be the same for all keys, so it plays no role, but we include it in the derivations for the sake of completeness.)  Denoting the parameters collectively by $\theta = \lbrace \pi, \beta, \nu \rbrace$, the probability within the model of a sequence of keys~$K_i = ( k^i_1,\ldots,k^i_{M_i} )$ for the $M_i$ steps of work~$i$~is
\begin{equation}
P(K_i|\theta) = P(k^i_1|\theta) \prod_{t=2}^{M_i} P(k^i_t|k^i_{t-1},\theta)
  = \pi_{k^i_1} \prod_{t=2}^{M_i} \beta_{k^i_{t-1}k^i_t},
\end{equation}
and the probability of the chords given the sequence of keys is
\begin{equation}
P(C_i|K_i,\theta) = \prod_{t=1}^{M_i} P(c^i_t|k^i_t)
  = \prod_{t=1}^{M_i} \nu_{k^i_tc^i_t}.
\end{equation}
Hence the probability of both the keys and the chords is
\begin{equation}
P(K_i,C_i|\theta) = P(C_i|K_i,\theta) P(K_i|\theta)
  = \pi_{k^i_1} \nu_{k^i_1c^i_1} \!\prod_{t=2}^{M_i} \beta_{k^i_{t-1}k^i_t} \nu_{k^i_tc^i_t}
\end{equation}
and the probability for the entire corpus, with keys $\cK = \lbrace K_1,\ldots, K_N \rbrace$ and chords $\cC = \lbrace C_1,\ldots, C_N \rbrace$, is the product of this expression over all works:
\begin{equation}
P(\cK,\cC|\theta) = \prod_{i=1}^N P(K_i,C_i|\theta)
  = \prod_{i=1}^N \pi_{k^i_1} \nu_{k^i_1c^i_1} \prod_{t=2}^{M_i} \beta_{k^i_{t-1}k^i_t} \nu_{k^i_tc^i_t}.
\label{eq:PKC}
\end{equation}

Using Bayes rule we write the probability of the unknowns $\cK$ and $\theta$ as
\begin{equation}
P(\cK,\theta|\cC) = P(\cK,\cC|\theta) {P(\theta)\over P(\cC)},
\end{equation}
which for given chords~$\cC$ is proportional to $P(\cK,\cC|\theta)$ assuming a uniform prior $P(\theta)=\mbox{constant}$ on the parameters~$\theta$.

We find our best estimate of the parameter values~$\theta$ by maximizing the log-probability
\begin{equation}
\log P(\theta|\cC) = \log \sum_\cK P(\cK,\theta|\cC) \propto \log \sum_\cK P(\cK,\cC|\theta).
\label{eq:em1}
\end{equation}
We now apply Jensen's inequality, which says that for any set of positive quantities $x_i,q_i>0$ with $i=1\ldots n$ and $\sum_{i=1}^n q_i = 1$ we have
\begin{equation}
\log \sum_i x_i \ge \sum_i q_i \log {x_i\over q_i}.
\end{equation}
Applied to Eq.~\eqref{eq:em1} this tells us that
\begin{equation}
\log \sum_\cK P(\cK,\cC|\theta) \ge \sum_\cK q(\cK) \log {P(\cK,\cC|\theta)\over q(\cK)},
\label{eq:em2}
\end{equation}
where $q(\cK)$ is any probability distribution over~$\cK$ satisfying $\sum_\cK q(\cK) = 1$.  The exact equality is achieved when
\begin{equation}
q(\cK) = {P(\cK,\cC|\theta)\over\sum_\cK P(\cK,\cC|\theta)},
\label{eq:estep1}
\end{equation}
which necessarily maximizes the right-hand side of~\eqref{eq:em2}, and a subsequent maximization with respect to~$\theta$ will then find the optimal parameter values.  Put another way, a double maximization of the right-hand side gets us the answer we want.  In the EM approach we perform this double maximization by simply maximizing with respect to $q(\cK)$ with $\theta$ held constant, then with respect to~$\theta$ with~$q(\cK)$ constant, and repeating until we reach convergence.  Maximization with respect to~$q(\cK)$ is achieved by applying Eq.~\eqref{eq:estep1} as we have said.  Maximization with respect to~$\theta$ is achieved by differentiating the right-hand side of~\eqref{eq:em2} and setting the result to zero.

For instance, the derivative with respect to~$\beta_{kk'}$ takes the form
\begin{align}
    {\partial\over\partial\beta_{kk'}} \biggl[ \sum_\cK q(\cK) \log P(\cK,\cC|\theta) - \lambda \sum_l \beta_{kl} \biggr] = 0,
\end{align}
where we have enforced the necessary constraint $\sum_l \beta_{kl} = 1$ with a Lagrange multiplier~$\lambda$.  Substituting from Eq.~\eqref{eq:PKC} for $P(\cK,\cC|\theta)$, taking the log, and performing the derivative, we then find that
\begin{equation}
\beta_{kk'} = {\sum_{i=1}^N \sum_{t=1}^{M_i-1} q^i_{t,t+1}(k,k')\over
                \sum_{k'} \sum_{i=1}^N \sum_{t=1}^{M_i-1} q^i_{t,t+1}(k,k')},
\label{eq:mstep2a}
\end{equation}
with
\begin{equation}
q^i_{t,t+1}(k,k') = \sum_\cK q(\cK) \delta_{k^i_t,k} \delta_{k^i_{t+1},k'},
\label{eq:qit1}
\end{equation}
where $\delta_{ij}$ is the Kronecker delta, which is 1 when $i=j$ and 0 otherwise.  A similar derivative with respect to $\nu_{kc}$, enforcing the constraint $\sum_c \nu_{kc} = 1$ with another Lagrange multiplier, gives
\begin{equation}
\nu_{kc} = {\sum_{i=1}^N \sum_{t=1}^{M_i} q^i_t(k) \delta_{c,c^i_t}\over \sum_{i=1}^N \sum_{t=1}^{M_i} q^i_t(k)},
\label{eq:mstep2b}
\end{equation}
where
\begin{equation}
q^i_{t,t+1}(k,k') = \sum_\cK q(\cK) \delta_{k^i_t,k} \delta_{k^i_{t+1},k'}.
\label{eq:qit2}
\end{equation}

The EM algorithm involves choosing an initial set of parameter values~$\theta = \lbrace \pi, \beta, \nu \rbrace$ and applying Eq.~\eqref{eq:estep1} to calculate~$q(\cK)$, then Eqs.~\eqref{eq:mstep2a}, \eqref{eq:mstep2b}, \eqref{eq:qit1}, and~\eqref{eq:qit2} to calculate new values of the parameters and iterating until the parameter values converge.  It can be proved that this procedure always converges to a \emph{local} optimum of the parameter values, but not to the global optimum.  In practice, therefore, one typically runs the algorithm several times and, if different runs converge to different optima, accepts the solution with the highest log-probability, Eq.~\eqref{eq:em1}.

Equations~\eqref{eq:mstep2a} and~\eqref{eq:mstep2b} are easily implemented.  The tricky part of the calculation is evaluating the probability distribution~$q(\cK)$ of Eq.~\eqref{eq:estep1} and from it the quantities $q^i_t(k)$ and $q^i_{t,t+1}(k,k')$.  A naive version of this calculation would involve performing the sum in the denominator of~\eqref{eq:estep1}, which would be computationally exorbitant, but luckily there exists an efficient algorithm for computing $q^i_t(k)$ and $q^i_{t,t+1}(k,k')$ directly, the \emph{forward-backward algorithm}.

\subsection{The forward-backward algorithm}
The quantity~$q^i_t(k)$ defined in Eq.~\eqref{eq:qit1} is equal to the probability that the prevailing key is~$k$ at step~$t$ of work~$i$.  Similarly $q^i_{t,t+1}(k,k')$ is the probability that the keys at consecutive steps $t$ and $t+1$ are $k$ and~$k'$.  The forward-backward algorithm computes these probabilities using a dynamic programming approach.  Focusing on a single work~$i$, we write the joint probability of the key and chord at the first step as
\begin{equation}
P(k_1,c_1|\theta) = P(k_1|\theta) P(c_1|k_1,\theta) = \pi_{k_1} \nu_{k_1c_1},
\label{eq:bwstart}
\end{equation}
where we temporarily suppress the superscript~$i$ for the sake of clarity, since it will be the same throughout.

Given this starting point, we can now calculate the set of subsequent probabilities
\begin{equation}
P(k_t,c_1\ldots c_t) = \sum_{k_{t-1}} P(k_{t-1},c_1\ldots c_{t-1})
  P(k_t|k_{t-1}) P(c_t|k_t),
\end{equation}
where we have also suppressed~$\theta$.  Defining the shorthand $a_t(k_t) = P(k_t,c_1\ldots c_t)$, this can then be written as
\begin{equation}
a_t(k) = \nu_{kc_t} \sum_{k'} \beta_{k'k} a_{t-1}(k').
\label{eq:fba}
\end{equation}
Equation~\eqref{eq:bwstart} tells us that $a_1(k) = \pi_k \nu_{kc_1}$, then applying~\eqref{eq:fba} repeatedly allows us to quickly calculate $a_t(k)$ for all keys~$k$ and steps~$t$.  This is the ``forward'' half of the forward-backward algorithm.

The backward half defines a complementary set of quantities $b_t(k_t) = P(c_{t+1}\ldots c_M|k_t)$, where $M$ is the total number of steps in the work.  We write
\begin{equation}
P(c_t\ldots c_M|k_{t-1}) = \!\sum_{k_t} P(c_{t+1}\ldots c_M|k_t) P(k_t|k_{t-1}) P(c_t|k_t),
\end{equation}
which can be written in the form
\begin{equation}
b_{t-1}(k) = \sum_{k'} \nu_{k'c_t} \beta_{kk'} b_t(k'),
\label{eq:fbb}
\end{equation}
from which, starting with $b_M(k)=1$ for all~$k$, we can quickly calculate all~$b_t(k)$.

Given the complete set of $a_t(k)$ and $b_t(k)$, we now have
\begin{equation}
P(k_t,C) = P(k_t,c_1\dots c_t) P(c_{t+1}\ldots c_M|k_t) = a_t(k_t) b_t(k_t),
\end{equation}
where $C = ( c_1,\ldots,c_M )$ as previously.  Thus
\begin{equation}
P(k_t|C) = {P(k_t,C)\over P(C)} = {P(k_t,C)\over\sum_{k_t} P(k_t,C)},
\end{equation}
or equivalently
\begin{equation}
q^i_t(k) = {a_t(k) b_t(k)\over\sum_k a_t(k) b_t(k)}.
\label{eq:fbqit1}
\end{equation}
Repeating this calculation for each work~$i$ we then have the full set of probabilities~$q^i_t(k)$.

Similarly we can write
\begin{align}
P(k_t,k_{t+1},C) &= P(k_t,c_1\dots c_t) P(c_{t+2}\ldots c_M|k_{t+1}) \nonumber\\
  &\hspace{4em}{}\times P(k_{t+1}|k_t) P(c_{t+1}|k_{t+1}) \nonumber\\
  &= a_t(k_t) b_{t+1}(k_{t+1})\,\beta_{k_tk_{t+1}} \nu_{k_{t+1}c_{t+1}},
\end{align}
so that
\begin{equation}
P(k_t,k_{t+1}|C) = {P(k_t,k_{t+1},C)\over P(C)}
  = {P(k_t,k_{t+1},C)\over\sum_{k_t,k_{t+1}} P(k_t,k_{t+1},C)},
\end{equation}
or
\begin{equation}
q_{t,t+1}(k,k') = {a_t(k) b_{t+1}(k')\,\beta_{kk'} \nu_{k'c_{t+1}}\over
                  \sum_{kk'} a_t(k) b_{t+1}(k')\,\beta_{kk'} \nu_{k'c_{t+1}}}.
\label{eq:fbqit2}
\end{equation}
Between them, Eqs.~\eqref{eq:fbqit1} and~\eqref{eq:fbqit2} now allow us to complete our EM algorithm and calculate the optimal values of the parameters.

\section{The Viterbi algorithm}
\label{sec:viterbi}
Using the Baum-Welch algorithm we can calculate the optimal values of the probability parameters $\beta_{kk'}$ and~$\nu_{kc}$.  Given these values we can then calculate the best estimate of the hidden variables using the Viterbi algorithm.  Again we use the language of keys and chords, in which case the hidden variables are the keys.

We define $x_t(k_t)$ to be the probability that the key is~$k_t$ on step~$t$ and the chords up to that point are~$c_1\ldots c_t$, given that the preceding keys $k_1\ldots k_{t-1}$ take their \emph{most likely} (i.e.~highest probability) values and given the values of the parameters~$\theta$.  That is,
\begin{equation}
x_t(k_t) = \max_{k_1\ldots k_{t-1}} P(k_1\ldots k_t,c_1\ldots c_t|\theta).
\end{equation}
We will henceforth suppress~$\theta$, since the parameters are fixed throughout this part of the calculation.

Now we note that
\begin{align}
& P(k_1\ldots k_t,c_1\ldots c_t) \nonumber\\
  &\qquad{} = P(k_1\ldots k_{t-1},c_1\ldots c_{t-1}) P(k_t|k_{t-1}) P(c_t|k_t),
\end{align}
and hence
\begin{align}
& \max_{k_1\ldots k_{t-1}} P(k_1\ldots k_t,c_1\ldots c_t|\theta) \nonumber\\
  &{} = \max_{k_{t-1}} \max_{k_1\ldots k_{t-2}}
           P(k_1\ldots k_{t-1},c_1\ldots c_{t-1}) P(k_t|k_{t-1}) P(c_t|k_t),
\end{align}
or
\begin{equation}
x_t(k) = \nu_{kc_t} \max_{k'} \beta_{k'k}\,x_{t-1}(k').
\label{eq:viterbi1}
\end{equation}
Starting from $x_1(k) = \pi_k \nu_{kc_1}$, we can use Eq.~\eqref{eq:viterbi1} to calculate the full set of quantities~$x_t(k)$ for all $t$ and~$k$.

The last of these quantities gives us the probability of $k_M$ when all other keys take their most likely values, so the most likely value of $k_M$ is simply
\begin{equation}
k_M = \argmax_k x_M(k),
\end{equation}
where $\argmax_k x(k)$ indicates the value of $k$ that gives the largest value of~$x$.  Then the most likely values of the remaining keys can be calculated in reverse order from
\begin{equation}
k_t = \argmax_k \beta_{kk_{t+1}} x_t(k).
\label{eq:viterbi2}
\end{equation}
The end result is the complete sequence of keys $k_1\ldots k_M$ that has the highest probability for the given values of the probability parameters.

\section{Implementation and running time}
\label{sec:implementation}
Implementation of the Baum-Welch algorithm is a straightforward numerical realization of the equations.  We choose random initial values for each of the probability parameters $\beta_{kk'}$ and~$\nu_{kc}$, and then for each work we repeatedly apply Eq.~\eqref{eq:fba} to calculate the quantities~$a_t(k)$, Eq.~\eqref{eq:fbb} to calculate the~$b_t(k)$, and then Eqs.~\eqref{eq:fbqit1} and~\eqref{eq:fbqit2} to calculate $q^i_t(k)$ and~$q^i_{t,t+1}(k,k')$.  Armed with these we then use Eqs.~\eqref{eq:mstep2a} and~\eqref{eq:mstep2b} to calculate updated values of $\beta_{kk'}$ and~$\nu_{kc}$, repeating the whole process until the values converge within a predetermined tolerance.

A technical issue arises with applications to longer works because the quantities~$a_t(k)$ and~$b_k(t)$ can become very small and underflow the numerical resolution of the computer.  To avoid this, we normalize these quantities at each step within a work.  For instance, after calculating $a_t(k)$ in Eq.~\eqref{eq:fba} we compute a normalized version
\begin{equation}
\tilde{a}_t(k) = {a_t(k)\over A_t},
\label{eq:tildea}
\end{equation}
with
\begin{equation}
A_t = \sum_k a_t(k),
\label{eq:Ak}
\end{equation}
then use this version in the next step of Eq.~\eqref{eq:fba}.  We also perform a similar normalization of~$b_k(t)$.  These changes have no effect on Eqs.~\eqref{eq:fbqit1} and~\eqref{eq:fbqit2} since any overall rescaling of $a_t(k)$ or $b_t(k)$ cancels in numerator and denominator of both equations, but they prevent underflows by ensuring that $a_t(k)$ and $b_t(k)$ both sum to~1, so that their average values cannot become arbitrarily small even for long works.

Implementation of the Viterbi algorithm is a similarly straightforward realization of Eq.~\eqref{eq:viterbi1}.  No iteration is needed here: one applies the algorithm just once to compute the optimal key sequence for each work.  Potential underflows can be prevented by normalizing the values of~$x_t(k)$ while having no effect on the keys computed in Eq.~\eqref{eq:viterbi2}.

Most of the running time is taken up by the calculation of the optimal parameter values using the Baum-Welch algorithm, and most of that time is taken by the forward-backward algorithm.  If there are $R$ distinct keys (where $R=24$ in our case) and if the total number of steps in all works in the corpus is $T = \sum_{i=1}^N M_i$, then the calculation of $a_t(k)$ from Eq.~\eqref{eq:fba} for all~$t$ and~$k$ takes time~$\Ord(TR^2)$, and similarly for the calculation of~$b_k(t)$.  This time gets multiplied by the number of iterations of the Baum-Welch algorithm needed to achieve convergence.  In our calculations we iterate until the values of the parameters change by less than one part in 10~million on each iteration, and this point is typically reached after about 100 iterations.  Total running time in practice for the full corpus of Bach chorales is about five minutes on a standard laptop (\textit{circa}~2024).

For practical applications to finding keys and chords in written music, however, one would not need to run the Baum-Welch algorithm at all.  In such applications one would normally pre-compute the optimal values of the probability parameters then apply those to infer keys and chords for any given work.  This would require running only the Viterbi algorithm for each work of interest, which is very fast, taking only a fraction of a second on a typical computer.

\section{Calculation of the data likelihood}
\label{sec:likelihood}
The Baum-Welch algorithm maximizes the probability of the model parameters given the data, Eq.~\eqref{eq:em1}, but is not guaranteed to find a \emph{global} maximum.  Like all expectation-maximization algorithms the Baum-Welch algorithm may erroneously converge to a local maximum of the probability.  To guard against this possibility we perform multiple runs of the algorithm with different random initial conditions and then use the results from the run that gives the highest probability.  This does not guarantee that we find the global maximum, but in practice we see the same best result occurring repeatedly, which strongly suggests that there is no better result to be found.

Calculating the value of the posterior probability $P(\theta|\mathcal{C})$ is equivalent to calculating the data likelihood $P(\mathcal{C}|\theta)$, since we assume a uniform prior on the parameters~$\theta$, and the likelihood can be calculated conveniently from results we already have.  As described in Appendix~\ref{sec:implementation}, we calculate as part of the Baum-Welch algorithm the quantities~$\tilde{a}_t(k)$ defined in Eq.~\eqref{eq:tildea}.  Applying this equation repeatedly we have
\begin{equation}
a_M(k_M) = P(k_M,c_1\ldots c_M|\theta) = \tilde{a}_M(k_M) \prod_{t=1}^M A_t
\end{equation}
with $A_t$ as in Eq.~\eqref{eq:Ak} and $M$ being the total number of steps in the work.  Summing over~$k_M$, we then have
\begin{align}
P(c_1\ldots c_M|\theta) &= \sum_{k_M} P(k_M,c_1\ldots c_t|\theta) \nonumber\\
  &= \sum_{k_M} \tilde{a}_M(k_M) \prod_{t=1}^M A_t
  = \prod_{t=1}^M A_t\,,
\label{eq:Pc}
\end{align}
where we have made use of the fact that $\sum_{k_M} \tilde{a}_M(k_M) = 1$ because of the normalization in Eq.~\eqref{eq:tildea}.  The total likelihood $P(\mathcal{C}|\theta)$ of all works in the entire data set is then the product of this quantity over all works.

\end{document}